\documentclass[journal]{IEEEtran}
\usepackage{times}
\usepackage{epsfig}
\usepackage{graphicx}
\usepackage{amsmath}
\usepackage{amssymb}

\usepackage{verbatim}
\usepackage{multirow}
\usepackage{tikz}
\newcommand*{\circled}[1]{\lower.7ex\hbox{\tikz\draw (0pt, 0pt)%
		circle (.5em) node {\makebox[1em][c]{\small #1}};}}

\ifCLASSINFOpdf
\else
\fi
\begin{document}

\title{Improving Description-based Person Re-identification by Multi-granularity \\Image-text Alignments}

\author{Kai~Niu,
	Yan~Huang,
	Wanli~Ouyang,~\IEEEmembership{Senior Member,~IEEE},
	and~Liang~Wang,~\IEEEmembership{Fellow,~IEEE}
	\thanks{K. Niu, Y. Huang and L. Wang are with the Center for Research on Intelligent Perception and Computing (CRIPAC), National Laboratory of Pattern Recognition (NLPR), Institute of Automation, Chinese Academy of Sciences (CASIA), Beijing 100190, China and, the University of Chinese Academy of Sciences (UCAS), Beijing 100049, China. L. Wang is also with the Center for Excellence in Brain Science and Intelligence Technology (CEBSIT), Institute of Automation, Chinese Academy of Sciences (CASIA), Beijing 100190, China. (e-mail: kai.niu@cripac.ia.ac.cn, yhuang@nlpr.ia.ac.cn, wangliang@nlpr.ia.ac.cn)}
	\thanks{W. Ouyang is with the School of Electrical and Information Engineering, University of Sydney, Sydney, NSW 2008, Australia. (e-mail: wanli.ouyang@sydney.edu.au)}
}

\maketitle

\begin{abstract}

Description-based person re-identification (Re-id) is an important task in video surveillance that requires discriminative cross-modal representations to distinguish different people. 
It is difficult to directly measure the similarity between images and descriptions due to the modality heterogeneity (the cross-modal problem).
And all samples belonging to a single category (the fine-grained problem) makes this task even harder than the conventional image-description matching task. 
In this paper, we propose a Multi-granularity Image-text Alignments (MIA) model to alleviate the cross-modal fine-grained problem for better similarity evaluation in description-based person Re-id. 
Specifically, three different granularities, \textit{i.e.}, global-global, global-local and local-local alignments are carried out hierarchically.
Firstly, the global-global alignment in the Global Contrast (GC) module is for matching the global contexts of images and descriptions.
Secondly, the global-local alignment employs the potential relations between local components and global contexts to highlight the distinguishable components while eliminating the uninvolved ones adaptively in the Relation-guided Global-local Alignment (RGA) module.
Thirdly, as for the local-local alignment, we match visual human parts with noun phrases in the Bi-directional Fine-grained Matching (BFM) module.
The whole network combining multiple granularities can be end-to-end trained without complex pre-processing.
To address the difficulties in training the combination of multiple granularities, an effective step training strategy is proposed to train these granularities step-by-step.
Extensive experiments and analysis have shown that our method obtains the state-of-the-art performance on the CUHK-PEDES dataset and outperforms the previous methods by a significant margin.

\end{abstract}


\IEEEpeerreviewmaketitle

\section{Introduction}
\begin{figure}[t]
	\centering
	\includegraphics[width=6cm]{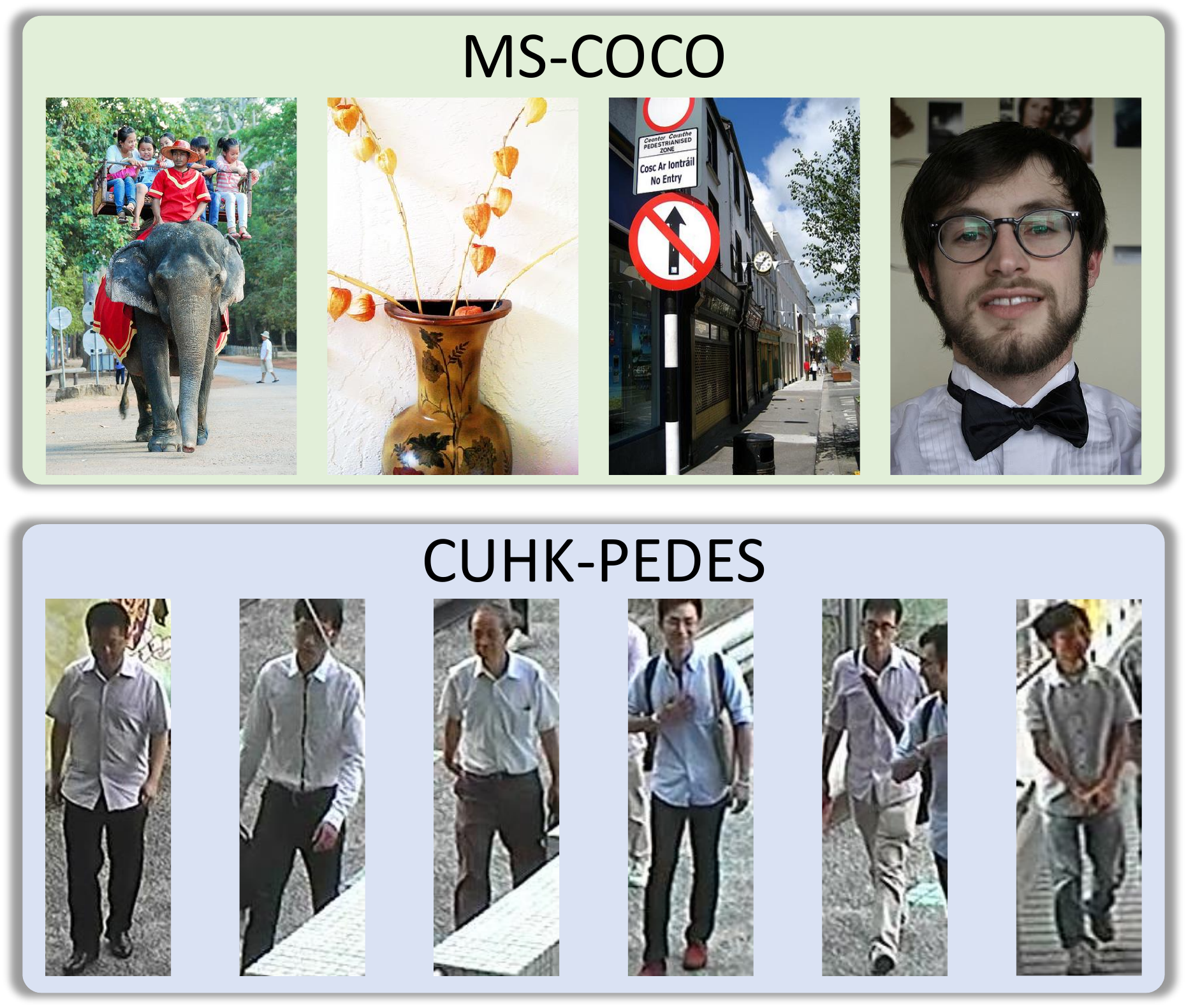}
	\vspace{-2mm}	
	\caption{The fine-grained problem of description-based person re-identification. Images in description-based person Re-id (CUHK-PEDES \cite{Person_Search_GNA-RNN} dataset, six different people) are much less distinguishable than the ones in image-description matching task (MS-COCO \cite{Karpathy&FeiFei} dataset), because they all belong to the same category, $i.e.$, pedestrian category. (Best viewed in colors.)} 
	\vspace{-3mm}
	\label{fig_problem}	
\end{figure}
\begin{figure*}[t]
	\centering
	\includegraphics[width=15cm]{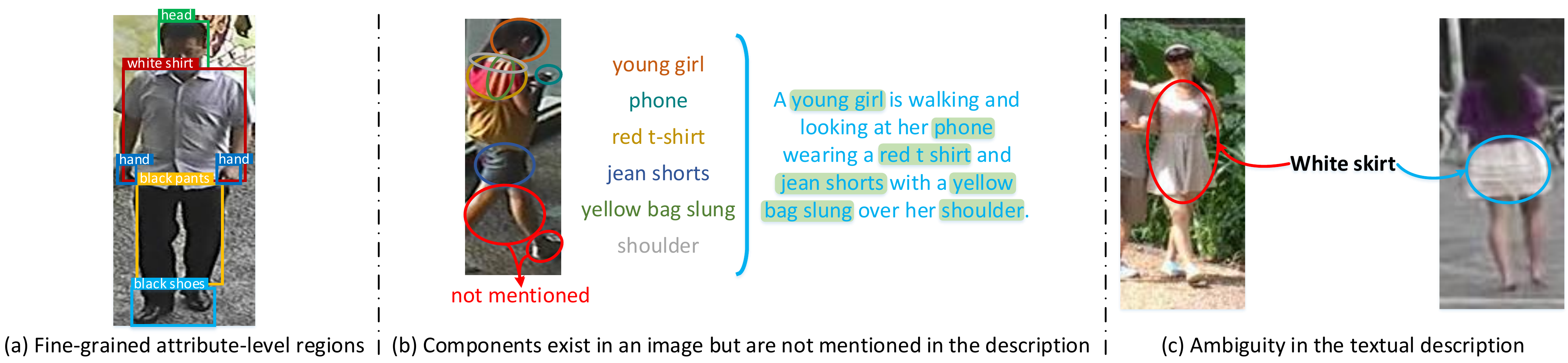}
	\vspace{-2mm}
	\caption{(a) Illustration of the fine-grained attribute-level regions in description-based person Re-id. (b) Illustration of the uninvolved components in image-sentence pair. The legs and shoes are not mentioned in the description and should not contribute to the visual representation. (c) Ambiguity due to the modality heterogeneity when using textual words to retrieve the matched image components. The `white skirt' can refer to several visual components from different people which contain skirts in different styles and sizes, covering different regions and parts of human body. (Best viewed in colors.)} 
	\label{fig_uninvolved}	
	\vspace{-2mm}	
\end{figure*}

\IEEEPARstart{P}{erson} re-identification (Re-id) \cite{PCB, Mask_guided, Person_Search_GNA-RNN, PDC, cheng2016person, cho2018pamm, yao2019deep, dai2019video} is an important task in video surveillance \cite{haritaoglu2000w, hu2007semantic, zhang2014background, lin2012integrating}, and research in this field can help fast and accurate retrieval in big visual data. 
An emerging task is to retrieve images of people across non-overlapping cameras with textual descriptions, which is named as description-based person Re-id.
Textual descriptions are easy to generate and can provide adequate and comprehensive information including semantic components and their potential relations to retrieve the matched person. 
Although studied from various perspectives \cite{Person_Search_GNA-RNN, Dual_Path, GLIA, PWM-ATH, IATVM}, there are still challenging problems that need to be better addressed.

Description-based person Re-id is a challenging task because the existing modality heterogeneity makes it difficult to directly measure the cross-modal similarity between images and descriptions. 
Although the conventional image and description matching problem has been widely studied \cite{sm_LSTM, SCAN, DAN, LSCO, Zhang_2018_ECCV, xu2017learning}, there is a specific difference in the task of description-based person Re-id.
All images in this task belong to the same category, \textit{i.e.}, pedestrian category (the fine-grained problem), making this task harder than only dealing with the modality heterogeneity.
As shown in Figure \ref{fig_problem}, six different people have similar suits, and it is obviously harder to distinguish them compared with the images in the conventional image-description matching problem which have various topics, scenes, styles and so on.
And the hand-annotated descriptions are also similar among different people due to the same category. 
Therefore, it is more difficult to solve the cross-modal fine-grained problem in the description-based person Re-id. 

To address the fine-grained problem, one straightforward idea is to apply the existing fine-grained component matching methods \cite{sm_LSTM, SCAN, DAN} to the new scenario of description-based person Re-id to enhance the discrimination of different features.
But there are still some problems that have not been well settled.

First, methods \cite{zheng2019pose, liu2018pose} based on pre-processing with external cues ($e.g.$, pose) for local component extraction need to be further fine-tuned or even re-trained additionally beforehand on the dataset of pedestrian.
So they can provide more accurate components for the subsequent fine-grained matching in person identification.
Unfortunately, there is no annotation of body part or body segment in the dataset of description-based person Re-id, which makes the fine-tuning or re-training impossible.
As for the region-based methods \cite{Bottom-up_Top-down_Attention, SCAN}, they need attribute-level annotations (as shown in Figure \ref{fig_uninvolved} (a)) to fine-tune the region proposal generation approaches for pre-processing due to the the fine-grained problem in person Re-id.
But attribute-based annotations are also not available in the pedestrian dataset.
And complex pre-processing approaches may lead to some difficulties in end-to-end training. 

Second, fine-grained part-based methods \cite{zhao2017deeply, PCB, yao2019deep} have shown great strength in the conventional image-based person Re-id, which do not require additional part labeling annotations for pre-processing.
However, it is non-trivial to carry out part-based fine-grained matching in the description-based person Re-id, and no work has been investigated from this aspect to the best of our knowledge.
The difficulties compared with the conventional image-based person Re-id mainly lie in two aspects.
On the one hand, a single image part may correspond to multiple separate words in the description, as shown by the part `yellow bag slung' in Figure \ref{fig_uninvolved} (b).
Thus the simple textual partition, $i.e.$, naturally dividing the sentence to separate words, is not appropriate for image-description fine-grained matching.
On the other hand, there are also some ambiguities due to the modality heterogeneity when using textual words to retrieve the matched image part.
Specifically, as shown in Figure \ref{fig_uninvolved} (c), `white skirt' can refer to several visual components from different people which contain skirts in different styles and sizes, covering different regions and parts of human body. 
And these ambiguities in fine-grained matching may cause confusion and harm the retrieval accuracy in the description-based person Re-id.
Therefore, adaptive local component alignment is necessary for cross-modal fine-grained matching.

In addition, only employing fine-grained component matching is not enough because it neglects the potential relations between local components and global contexts, which can also be used to improve the cross-modal similarity evaluation.
To be more specific, the cross-modal global-local relations can be used as filters to eliminate the uninvolved components in the other modality in an image-sentence pair.
For example in Figure \ref{fig_uninvolved} (b), the legs and shoes are not mentioned in description, thus these attributes should not contribute to the visual semantic representations accordingly.
In this example, sentence description provides cross-modal mutual information that helps to ignore uninvolved visual cues, and this also applies when visual information is used for discarding uninvolved textual components.
Based on the relation-guided filtering process, we can obtain better aggregated representations for measuring more accurate cross-modal similarities.

\begin{figure}[t]
	\centering
	\includegraphics[width=8cm]{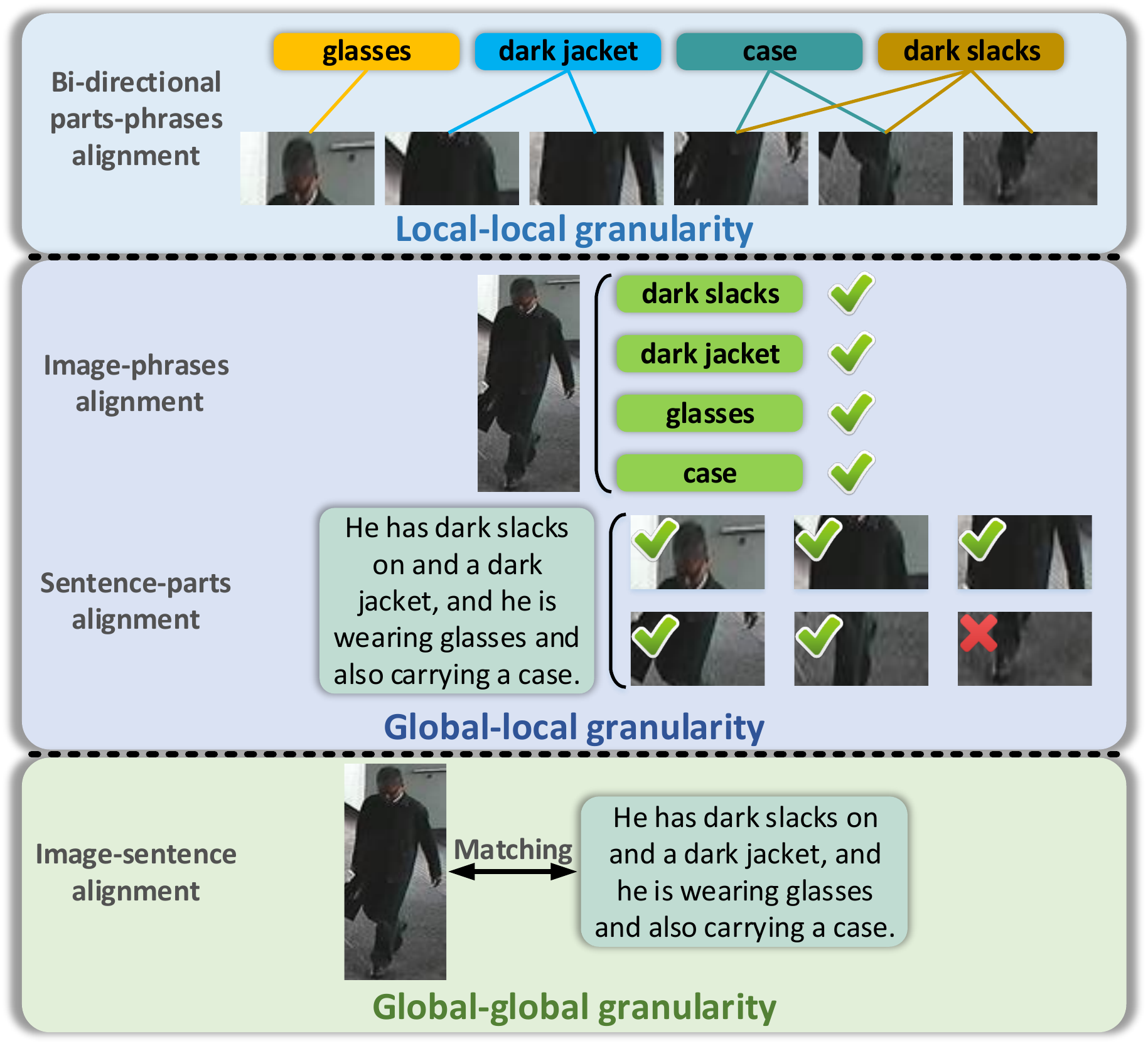}
	\vspace{-2mm}
	\caption{The multi-granularity image-text alignments. There are three different granularities in our solution, $i.e.$, global-global, global-local and local-local alignments. The tick icon in the global-local granularity means that the component is mentioned in global context in the other modality, and the cross icon means that it is not mentioned. For example, `shoes' are not mentioned in the corresponding description, thus the last image part is with a cross icon, which indicates that this image part is less important according to the textual description. (Best viewed in colors.)} 
	\vspace{-2mm}
	\label{fig_structure}		
\end{figure}

More than the fine-grained component matching and relation-guided matching that consider the fine-grained problem to enhance the discrimination of features, the global contexts are also important in person identification.
It is because that global contexts consist of more information including not only local components but their spatial relations (mainly in images) and orders (mainly in descriptions) comprehensively.
These potential semantic aspects also contribute to identifying a pedestrian more accurately.
On the whole, as explained in Figure \ref{fig_structure}, we consider the foregoing fine-grained component matching, relation-guided matching, and the global context matching as different granularities to carry out multi-granularity cross-modal alignments hierarchically.
The three granularities can complement each other and provide comprehensive cross-modal similarity evaluation. 

Although our method can be end-to-end trained, it does not mean that training all modules simultaneously is a good training strategy.
In fact, the combination of multiple granularities brings some difficulties in training.
For one thing, local components and global contexts lie in different semantic levels and there are some differences in the objectives to use in training.
To be more specific, global contexts contain more than local components but their potential dependency ($e.g.$, spatial relations in images and word orders in descriptions), so they have tighter relevance to person identity than local components.
And they are appropriate to be trained under the supervision of person identity more than only the cross-modal matching.
For another, local component extraction approaches may inevitably bring some ambiguities in fine-grained component representations.
For example, multiple attributes or incomplete attribute may be divided into a single image part, and this problem is likely to harm the global feature extraction when trained together.
Therefore, we empirically find that it is more effective to train the global contexts and the local components hierarchically and step-by-step.

In summary, this paper proposes a \textbf{Multi-granularity Image-text Alignments} (MIA) model to alleviate the cross-modal fine-grained problem for better similarity evaluation in tha task of description-based person Re-id.
Specifically, three different granularities, $i.e.$, global-global, global-local and local-local alignments are carried out hierarchically as shown in Figure \ref{fig_structure}.
Firstly, the global-global alignment in the Global Contrast (GC) module is for matching the global contexts of images and descriptions.
Secondly, the global-local alignment employs the potential relations between local components and global contexts to highlight the distinguishable components while eliminating the uninvolved ones adaptively in the Relation-guided Global-local Alignment (RGA) module.
Thirdly, as for the local-local alignment, we match visual human parts with noun phrases in the Bi-directional Fine-grained Matching (BFM) module.
And the whole network with multiple granularities can be end-to-end trained without complex pre-processing.
To address the difficulties in training the combination of multiple granularities, an effective step training strategy is proposed to train these granularities step-by-step.
We have obtained the state-of-the-art performance on the CUHK-PEDES \cite{Person_Search_GNA-RNN} dataset, and outperformed the previous methods by a significant margin.	
The main contributions are as follows:
\begin{itemize}
	\setlength{\itemsep}{0pt}
	\setlength{\parsep}{0pt}
	\setlength{\parskip}{1pt}
	\item To the best of our knowledge, we are probably the first to match visual human parts with noun phrases for description-based person Re-id.
	\item To alleviate the cross-modal fine-grained problem, we propose a multi-granularity image-text alignments model for description-based person Re-id.
	Three different granularities, \textit{i.e.}, global-global, global-local and local-local alignments are carried out hierarchically.
	They consider matching the global contexts, using global-local relations to filter the uninvolved components and the bi-directional fine-grained matching, respectively, for more accurate cross-modal matching.   
	And the proposed model is end-to-end trainable.
	\item To better train the combination of multiple granularities, an effective step training strategy is proposed to train the whole model step-by-step.
	\item We obtain the state-of-the-art performance on the CUHK-PEDES dataset, significantly outperforming other previous methods.
\end{itemize} 

\section{Related Work}
\begin{figure*}[t]
	\centering
	\includegraphics[width=17cm]{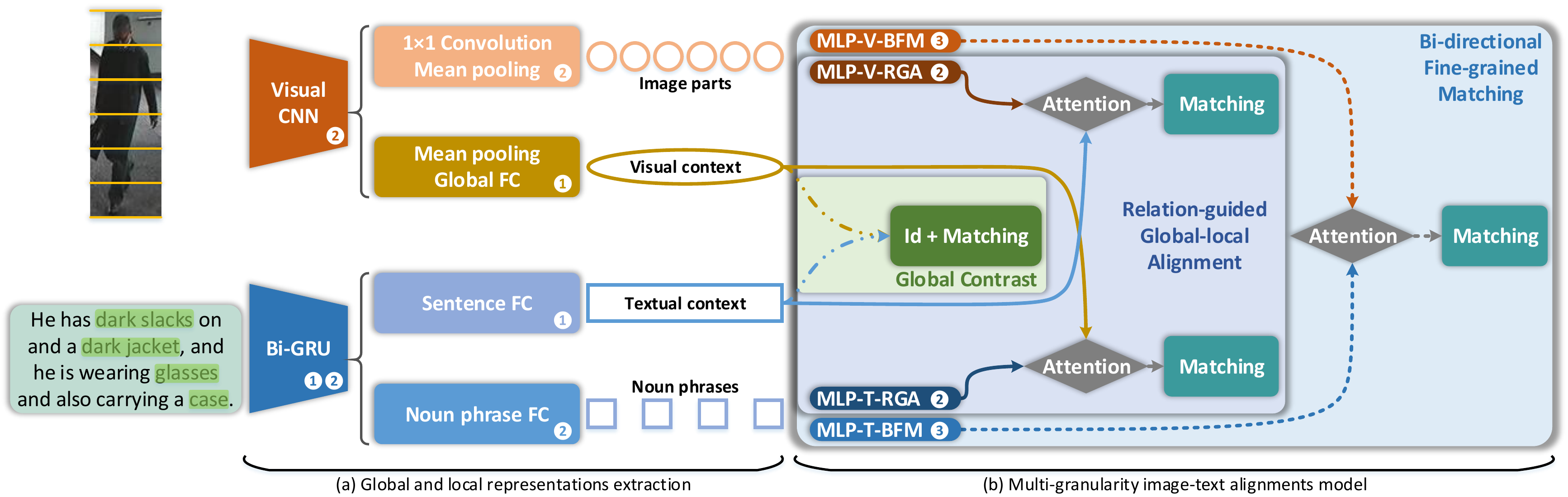}
	\vspace{-2mm}
	\caption{The overall framework of our solution. There are mainly two parts inside the framework: the (a) global and local representation extraction and the (b) multi-granularity image-text alignments model. The numbers on different blocks show the steps that they are trained in our step training strategy, respectively. And the proposed step training strategy will be explained in detail in the following Section \ref{sec_train_strategy}. (Best viewed in colors.)}	
	\label{fig_framework}
	\vspace{-2mm}	
\end{figure*}

\subsection{Visual-Semantic Embedding}

With the rapid growth of multi-modal data, the relations and alignments among different modalities have drawn much attention recently.
Frome $et \ al.$ \cite{VSE} propose the Visual Semantic Embedding (VSE) framework, which aligns global image features with sentence features by using ranking loss. 
Faghri $et \ al.$ \cite{VSE++} penalize the VSE model according to the hardest negative examples in the loss function and further improve the image-sentence alignment. 
Huang $et \ al.$ \cite{sm_LSTM} propose a selective multi-modal Long Short Term Memory network (sm-LSTM) for image-text matching.
The sm-LSTM includes a multi-modal context-modulated attention scheme which can selectively attend to a pair of instances of image and sentence at each timestep.	
Lee $et \ al.$ \cite{SCAN} propose the Stacked Cross Attention Network (SCAN), which discovers the cross-modal alignments by a fine-grained attention scheme on regions in image and words in sentence.
Beyond the fundamental image-text matching, there are more emerging and attractive applications related to visual-semantic embedding, such as image captioning \cite{Show_Attend_Tell, Self-critical-Captioning, Knowing-Captioning, Convolutional-Captioning} and visual question answering \cite{VQA, Hierarchical-VQA, Ask-attend-answer, Tips-Tricks-VQA, xue2017unifying}.
Anderson $et \ al.$ \cite{Bottom-up_Top-down_Attention} propose a combined bottom-up and top-down attention mechanism for image captioning and visual question answering. 
The bottom-up attention proposes features of salient image regions by Faster R-CNN \cite{Faster-R-CNN}, while the top-down attention determines feature weightings.

Unlike them, the fine-grained problem is the major difficulty in distinguishing different people in the description-based person Re-id, which needs to be carefully addressed.

\subsection{Person Re-identification}
Person re-identification has gained increasing attention from both academia and industry recently, and state-of-the-art approaches are mostly dominated by the emerging deep learning techniques.
Su $et \ al.$ \cite{PDC} propose a pose-driven deep convolutional model to learn improved features, which leverages the human parts to alleviate the pose variations and learn robust feature representations from both the global image and different local parts.
Liu $et \ al.$ \cite{liu2018pose} propose a pose-transferrable person Re-id framework which utilizes pose transferred sample augmentations (with identity supervision) to enhance Re-id model training.
Song $et \ al.$ \cite{Mask_guided} introduce the binary segmentation masks and further design a mask-guided contrastive attention model to learn features separately from the body and
background regions with region-level triplet loss.

Different from the previous approaches that exploit additional uni-modal cues, $e.g.$, pose and mask, for discriminative representation learning, the most severe difficulty in description-based person Re-id is the modality heterogeneity.
And it is non-trivial to address the cross-modal fine-grained problem in the description-base person Re-id.

\subsection{Description-based Person Re-identification}
Li $et \ al.$ \cite{Person_Search_GNA-RNN} propose the first large-scale person description dataset, CUHK PErson DEScription dataset (CUHK-PEDES), which contains person images with detailed natural language annotations. 
They also provide the Recurrent Neural Network with Gated Neural Attention mechanism model (GNA-RNN) with unit-level attentions and word-level gates to determine the cross-modal affinity with only matching objective.
After that, Li $et \ al.$ \cite{IATVM} further propose an identity-aware two-stage framework for textual-visual matching. 
They first adopt Cross-Modal Cross-Entropy loss (CMCE) in stage-1 for screening easy incorrect matchings, which uses person identity information as supervision.
Then they verify hard matchings with a co-attention mechanism, which jointly learns the visual spatial attention and latent semantic attention in stage-2. 
These two solutions mentioned above regard each hidden state from LSTM as word-level textual representation, which may incur some noise because a hidden state contains a complex semantic mixture of the current word and previous words. 
And they only pay attention to one single direction when using the fine-grained matching or attention scheme for representation enhancement, $i.e.$, only using text for weighting different visual components.
Chen $et \ al.$ \cite{GLIA} improve visual representations by global and local cross-modal associations. 
The global image-language association is established according to the identity labels, and the local association focuses on improving the visual representations by phrase reconstruction.

Different from them, we are probably the first to match visual human parts with noun phrases for description-based person Re-id, and further propose a multi-granularity image-description alignments model to carry out more accurate person identification. 

\section{Proposed Approach}		
The overall framework with our Multi-granularity Image-text Alignments (MIA) model is shown in Figure \ref{fig_framework}.
There are mainly two parts inside the framework: (a) global and local representations extraction and (b) multi-granularity image-text alignments model.

In part (a), we use convolutional neural networks (CNN) \cite{CNN,VGG,ResNet} to extract the visual feature maps in the image path.
Then the path is divided into two for global context features and image part features, respectively.
We use a global mean pooling layer and a fully connected layer (FC layer) in sequence to obtain the global visual representation. 
We employ $1 \times 1$ convolution and local mean pooling on respective image parts to obtain part features.
As for the textual path, sentence encoding and phrase encoding share the same bi-directional gated recurrent unit (Bi-GRU) \cite{GRU, chung2014empirical} and then have different FC layers.

For the MIA model in part (b), there are mainly three modules corresponding to the three granularities.
To be more specific, the Global Contrast (GC) module is used to carry out global-global alignment.
And it uses the global visual and textual context representations to obtain a fundamental global-global similarity.
The Relation-guided Global-local Alignment (RGA) module is for global-local relation filtering, which exploits cross-modal relation alignments to filter the uninvolved attributes for better aggregated representations.
And an intermediate global-local similarity is computed in the RGA module.
Then the Bi-directional Fine-grained Matching (BFM) module is for local-local alignment based on the trained fine-grained local components.
By combining these three modules for different granularities hierarchically, we can obtain more comprehensive cross-modal similarity evaluation.

\subsection{Multi-granularity Image-text Alignments (MIA)}

We present the details of the three different modules in our MIA model as follows.

\subsubsection{Global Contrast (GC)}
For the fundamental global-global granularity, we use the conventional CNNs for visual encoding and Bi-GRU for textual encoding.
For the image $I$, the feature maps from CNN are sequentially passed through a global mean pooling layer and a FC layer to obtain the global visual context representation, $\textbf{I} \in {\mathbb{R}^{V}}$.
As for the description $T$, we first embed each word ${\textbf{w}} \in {\mathbb{R}^{W}}$ inside $T$ to a vector $\textbf{x} \in {\mathbb{R}^{{E}}}$ by 
\begin{equation}\label{equ_text_embedding}
\begin{aligned}
{\textbf{x}} = {W_e} \times {\textbf{w}},
\end{aligned}
\end{equation} 
where ${W_e} \in {\mathbb{R}^{{E} \times {W}}}$.
Then we input all these vectors sequentially through a Bi-GRU,
\begin{equation}\label{equ_text_gru}
\begin{aligned}
\begin{array}{l}
\overrightarrow {{\textbf{h}_t}}  = \overrightarrow {GRU} ({\textbf{x}_t},\overrightarrow {{\textbf{h}_{t - 1}}} ) \\
\overleftarrow {{\textbf{h}_t}}  = \overleftarrow {GRU} ({\textbf{x}_t},\overleftarrow {{\textbf{h}_{t - 1}}} ).
\end{array} 
\end{aligned}
\vspace{-1mm}	
\end{equation} 
After that, we concatenate the forward hidden state $\overrightarrow {{\textbf{h}_F}}  \in {\mathbb{R}^{H}}$ and backward $\overleftarrow {{\textbf{h}_{F'}}}  \in {\mathbb{R}^{H}}$ at separate last time-step $F$ and $F'$.
And the sentence FC layer is employed for getting the final representation $\textbf{T} \in {\mathbb{R}^{C}}$ of the description $T$, as shown by
\begin{equation}\label{equ_text_concat_fc_global}
\begin{aligned}
{\textbf{T}} = {W_g} \times [\overrightarrow {{\textbf{h}_F}} ,\overleftarrow {{\textbf{h}_{F'}}} ] + {\textbf{b}_g},
\end{aligned}
\end{equation} 
where $W_{g} \in {\mathbb{R}^{{C} \times {2H}}} $ and $\textbf{b}_{g} \in {\mathbb{R}^{C}}$ are the parameters in the sentence FC layer for the description $T$.	
And $[ \cdot , \cdot ]$ means the concatenation of two vectors.
The global-global similarity between image $I$ and description $T$ is computed by 
\begin{equation}
	\begin{gathered}
	{s_G} = sim({\bf{I}},{\bf{T}}),
	\end{gathered}
	\vspace{-1mm}	
\end{equation} 
where $sim ( \cdot , \cdot )$ denotes the similarity function between the two feature vectors.

\subsubsection{Relation-guided Global-local Alignment (RGA)} \label{sec_rga}
For relational modeling, we first have to extract the fine-grained image parts and noun phrases.
Then the global-local relations can be modeled based on the fine-grained component features and global contexts. 
The approach in \cite{PCB} splits the feature maps from visual CNN into several non-overlapping parts equally along the vertical direction.
And it introduces almost no extra computational cost, better than the pose-based or region-based approaches which require complex pre-processing beforehand.
Therefore, visual partition can be done along with the training procedure and enables our method to be end-to-end trainable.
As for the textual description, noun phrases have more semantic concepts and more precise correspondences with image parts than separate words.
Therefore, we focus on noun phrases and encode them separately, different from \cite{SCAN} that models all words in sentence by their hidden states from LSTM.
The reason is that a hidden state contains a complex semantic mixture of the current word and previous words, which may bring noise in the following cross-modal matching.

After obtaining the visual human parts and textual noun phrases, directly using these fine-grained local components for cross-modal matching is inappropriate, because there exist some ambiguities due to the modality heterogeneity and imperfect partition approaches, and the local component representations are not well trained yet.
Therefore, we employ the attention mechanism to first carry out relation-guided global-local alignment to improve the quality of local component representations.
And an intermediate cross-modal similarity can be obtained based on the attentively aggregated representations and the global contexts.

\begin{figure}[t]
	\centering
	\includegraphics[width=8.5cm]{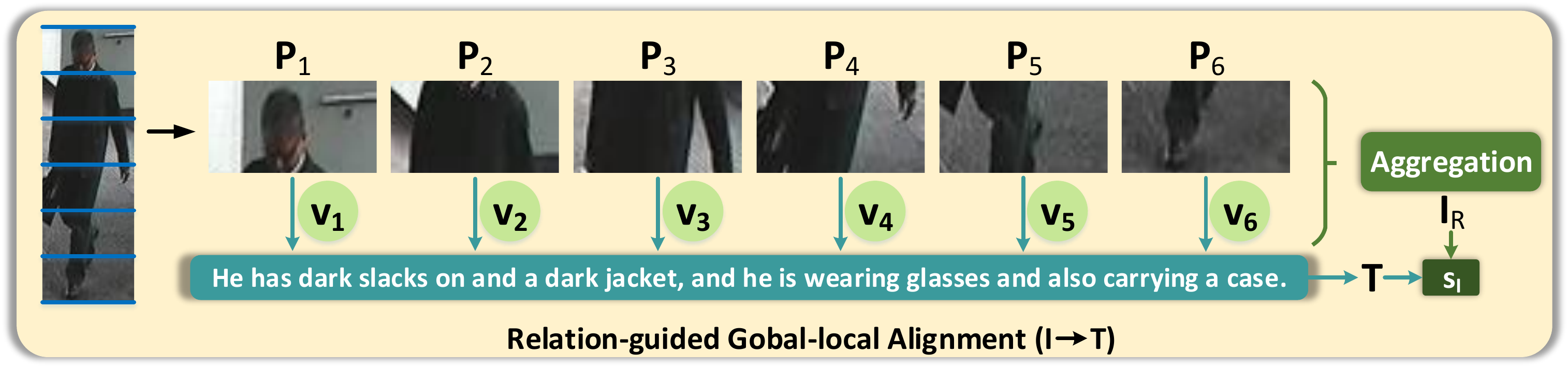}
	\vspace{-2mm}
	\caption{Illustration of obtaining the global-local similarity in the $I \to T$ direction in the RGA module. (Best viewed in colors.)}	
	\label{fig_relation}
	\vspace{-3mm}
\end{figure}

For the image $I$, we obtain $n$ local features corresponding to different non-overlapping image parts following \cite{PCB}, which are denoted as $\textbf{P}_1, \textbf{P}_2, ..., \textbf{P}_n \in {\mathbb{R}^{P}}$.
As for description $T$, we use the Natural Language ToolKit (NLTK) \cite{NLTK} for syntactic analysis, word segmentation and part-of-speech tagging, and then obtain several noun phrases. 
This extraction procedure can be handled dynamically along with the training procedure, which benefits the end-to-end training.
Similar to the whole description encoding, we use Equations \ref{equ_text_embedding} and \ref{equ_text_gru}, and employ another FC layer in 
\begin{equation}\label{equ_text_concat_fc_local}
\begin{aligned}
{\textbf{N}} = {W_l} \times [\overrightarrow {{\textbf{h}_F}} ,\overleftarrow {{\textbf{h}_{F'}}} ] + {\textbf{b}_l} 
\end{aligned}
\end{equation} 	
for getting the representation of a noun phrase. 
$W_{l} \in {\mathbb{R}^{{N} \times {2H}}} $ and $\textbf{b}_{l} \in {\mathbb{R}^{N}}$ are the parameters in the noun phrase FC layer.
We do not restrict the number $m$ of noun phrases extracted from a sentence, $i.e.$, $m$ is different for different descriptions, and obtain the features of $\textbf{N}_1, \textbf{N}_2, ..., \textbf{N}_m \in {\mathbb{R}^{N}}$.

Based on the image part representations $\textbf{P}_1, \textbf{P}_2, ..., \textbf{P}_n$ and noun phrase representations $\textbf{N}_1, \textbf{N}_2, ..., \textbf{N}_m$, we have two opposite directions in relation-guided global-local alignment, $i.e.$, image-guided phrase alignment ($T \to I$) and sentence-guided part alignment ($I \to T$).
Figure \ref{fig_relation} shows the $I \to T$ direction as an example.
We first employ a cross-modal attention method to determine the relations $v_i$ between all the image parts $\textbf{P}_1, \textbf{P}_2, ..., \textbf{P}_n$ and the global textual context $\textbf{T}$.
Specifically, each $v_i$ is computed by
\begin{equation} \label{equ_textual_saliency}
	\begin{gathered}
	{v_i} = \frac{{\exp (sim(ML{P_V}({{\bf{P}}_i}),{\bf{T}}))}}{{\sum\nolimits_{i = 1}^n {\exp (sim(ML{P_V}({{\bf{P}}_i}),{\bf{T}}))} }}.
	\end{gathered}	
	\vspace{-1mm}	
\end{equation}
$ML{P_V}(\cdot)$ means multi-layer perceptron for visual parts, $i.e.$, MLP-V-RGA in Figure \ref{fig_framework}, and $sim ( \cdot , \cdot )$ denotes the similarity function between $\textbf{P}_i$ and $\textbf{T}$.
Then we use 
\begin{equation}\label{equ_image_agg_visual_saliency}
	\begin{gathered}
	{{\bf{I}}_{R}} = \sum\nolimits_{i = 1}^n {{v_i} \cdot {{\bf{P}}_i}}
	\end{gathered}
	\vspace{-1mm}	
\end{equation} 	
to selectively aggregate the part representations $\textbf{P}_i$ to the relation-guided visual representation $\textbf{I}_{R} \in {\mathbb{R}^{P}}$.
This feature aggregation process is supervised by the global-local relation indicator $v_i$, which indicates the importance between different image parts and the whole description. 
After that, the intermediate cross-modal similarity in the $I \to T$ direction is
\begin{equation}
s_I = sim({{\bf{I}}_R},{\bf{T}}),
\vspace{-1mm}	
\end{equation}
which is regarded as the global-local similarity inside the intermediate hierarchy RGA of our MIA model.

Similarly, we can obtain the relation-guided textual representation $\textbf{T}_{R} \in {\mathbb{R}^{N}}$ in the opposite $T \to I$ direction by 
\begin{equation}\label{equ_visual_saliency}
	\begin{gathered}
	{t_j} = \frac{{\exp (sim(ML{P_T}({{\bf{N}}_j}),{\bf{I}}))}}{{\sum\nolimits_{j = 1}^m {\exp (sim(ML{P_T}({{\bf{N}}_j}),{\bf{I}}))} }},
	\end{gathered}	
\end{equation} 
\begin{equation}\label{equ_text_agg_visual_saliency}
	\begin{gathered}
	{{\bf{T}}_{R}} = \sum\nolimits_{j = 1}^m {{t_j} \cdot {{\bf{N}}_j}}.
	\end{gathered}		
\end{equation}	
$ML{P_T}(\cdot)$ means multi-layer perceptron for noun phrases, $i.e.$, MLP-T-RGA in Figure \ref{fig_framework}.
And the corresponding intermediate similarity after relation-guided global-local alignment is 
\begin{equation}
s_T = sim({{\bf{T}}_R},{\bf{I}}).
\end{equation}

\subsubsection{Bi-directional Fine-grained Matching (BFM)}
In the top level of our MIA model, we carry out bi-directional fine-grained matching between visual human parts and textual noun phrases for local-local granularity similarities.
The local feature extraction of image parts and noun phrases employs the modules that have been trained in the foregoing GC and RGA modules.
And another two MLPs (MLP-T-BFM and MLP-V-BFM in Figure \ref{fig_framework}) in the BFM module provide adaptation for appropriate parts-phrases matching.
Based on the component representations, the BFM module obtains distinguishable fine-grained cross-modal similarities and identifies a pedestrian having small differences with others.

\begin{figure}[t]
	\centering
	\includegraphics[width=8.5cm]{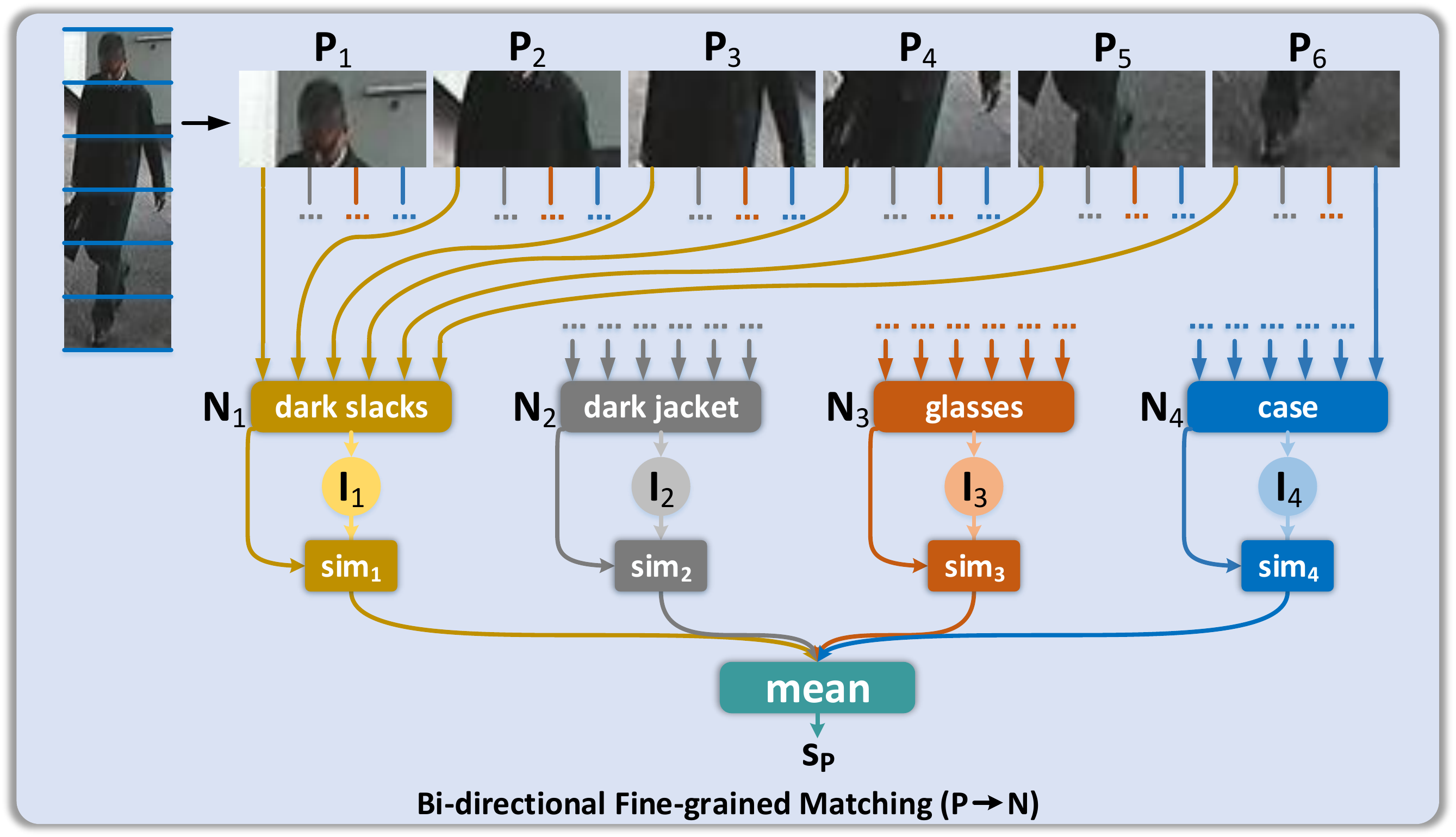}
	\vspace{-2mm}
	\caption{Illustration of obtaining the local-local similarity in the $P \to N$ direction in the BFM module. (Best viewed in colors.)}	
	\label{fig_matching}
	\vspace{-2mm}
\end{figure}

To better exploit cross-modal correspondences, we have two opposite directions in fine-grained matching, $i.e.$, noun-phrase-related direction ($P \to N$) and part-related direction ($N \to P$).
Figure \ref{fig_matching} shows an example of the $P \to N$ direction.
We first select a noun phrase, $e.g.$, $\textbf{N}_1$ of \emph{dark slack}, and evaluate the similarity between $\textbf{N}_1$ and all the image parts, $\textbf{P}_1, \textbf{P}_2, ..., \textbf{P}_n$. 
Then we refer to these similarity values and use an attention mechanism to adaptively obtain the combined single-noun-phrase-related visual representation, $\textbf{I}_1 \in {\mathbb{R}^{P}}$, as shown by the path in golden color in Figure \ref{fig_matching}.
Using the same steps for each noun-phrase feature $\textbf{N}_i$, we can obtain all the combined single-noun-phrase-related visual representations $\textbf{I}_1, \textbf{I}_2, ..., \textbf{I}_m \in {\mathbb{R}^{P}}$ by
\begin{equation} \label{equ_single-noun-phrase-related_visual_feat}
	\begin{gathered}
	{{\bf{I}}_i} = \sum\nolimits_{k = 1}^n {{\alpha _{i,k}} \cdot {{\bf{P}}_k}},
	\end{gathered}			
\end{equation} 		
	\begin{equation}\label{equ_single-noun-phrase-related_visual_alpha}
	\begin{gathered}
	{\alpha _{i,k}} = \frac{{\exp (sim(ML{P_T}({{\bf{N}}_i}),ML{P_V}({{\bf{P}}_k})))}}{{\sum\nolimits_{k = 1}^n {\exp (sim(ML{P_T}({{\bf{N}}_i}),ML{P_V}({{\bf{P}}_k})))} }}.
	\end{gathered}	
\end{equation}
$ML{P_T}(\cdot)$ means the multi-layer perceptron MLP-T-BFM for noun phrases, and $ML{P_V}(\cdot)$ indicates the MLP-V-BFM for visual parts, as shown in Figure \ref{fig_framework}.
$sim ( \cdot , \cdot )$ denotes the similarity function between the two feature vectors, and $\exp ( \cdot )$ means the exponential operation.
After obtaining $\textbf{I}_1, \textbf{I}_2, ..., \textbf{I}_m$, the local-local similarity in the $P \to N$ direction is
\begin{equation}
	\begin{gathered}
	{s_P} = \frac{1}{m}\sum\limits_{k = 1}^m {sim({{\bf{I}}_k},{{\bf{N}}_k})}.
	\end{gathered}
\end{equation} 

Similarly in the opposite $N \to P$ direction, we can obtain the combined single-image-part-related textual representations $\textbf{T}_1, \textbf{T}_2, ..., \textbf{T}_n \in {\mathbb{R}^{N}}$ by
\begin{equation} \label{equ_single-image-part-related_textual_feat}
	\begin{gathered}
	{{\bf{T}}_j} = \sum\nolimits_{k = 1}^m {{\beta _{j,k}} \cdot {{\bf{N}}_k}},
	\end{gathered}
\end{equation}	 	
\begin{equation}\label{equ_single-image-part-related_beta}
	\begin{gathered}
	{\beta _{j,k}} = \frac{{\exp (sim(ML{P_V}({{\bf{P}}_j}),ML{P_T}({{\bf{N}}_k})))}}{{\sum\nolimits_{k = 1}^m {\exp (sim(ML{P_V}({{\bf{P}}_j}),ML{P_T}({{\bf{N}}_k})))} }}.
	\end{gathered}
\end{equation} 	
And the local-local similarity in the $N \to P$ direction is 
\begin{equation}
	\begin{gathered}
	{s_N} = \frac{1}{n}\sum\limits_{k = 1}^n {sim({{\bf{T}}_k},{{\bf{P}}_k})}.
	\end{gathered}
\end{equation} 

\subsubsection{Similarity Fusion} \label{sec_fusion}
From the aforementioned multi-granularity alignments in separate modules, we obtain five similarities which can be divided into three different categories.
Specifically, $s_G$ from the GC module can be regarded as the global-global similarity, and $s_I$ and $s_T$ are the intermediate global-local similarities in the RGA module.
In the top-level BFM module, $s_P$ and $s_N$ are regarded as the local-local similarities.
To properly fuse these similarities, we introduce two hyper-parameters to adjust their proportions,
\begin{equation} \label{equ_sim_fusion}
	\begin{gathered}
	{s_F} = {s_G} + {\lambda _1} \cdot {s_R} + {\lambda _2} \cdot s{}_L, \\
	{s_R} = \left( {{s_I} + {s_T}} \right)/2, \\
	{s_L} = \left( {{s_P} + {s_N}} \right)/2,
	\end{gathered}
\end{equation} 
where $s_F$ indicates the final cross-modal similarity, and $s_R$ and $s_L$ indicate the similarities in RGA and BFM, respectively.

\subsection{Learning Procedure} \label{sec_learning_procedure}
We employ two kinds of objectives and propose an effective step training strategy.
In the following, we provide the details of the objectives and the training strategy.

\subsubsection{Objectives} \label{sec_objectives}
Two different objectives are used in training, $i.e.$, \textbf{Identity objective} and \textbf{Matching objective}.
The identity objective comes from that identities (ID) of pedestrians can be regarded as categories for classifying images and descriptions, and the matching objective is commonly used in the conventional cross-modal retrieval.

\textbf{Identity objective}: We regard different IDs in the training set as the number of categories, and classify images and descriptions to the corresponding ID category separately. 
The loss value of the identity objective is 
\begin{equation}\label{equ_identityloss}	
\begin{array}{l}
\begin{aligned}
{{\bf{P}}_I} & = softmax({W_{s}} \cdot \bf{I} + {\textbf{b}_{s}}),\\	
{L_{I}} & =  - log ({\textbf{P}_{I}}(ID)),\\
{{\bf{P}}_T} & = softmax({W_{s}} \cdot \bf{T} + {\textbf{b}_{s}}),\\
{L_{T}} & =  - log ({\textbf{P}_{T}}(ID)).
\end{aligned}
\end{array}			
\end{equation} 
${W_{s}} \in {\mathbb{R}^{{ID} \times {G}}}$ is the shared matrix for mapping the features of images and descriptions into the same space, whose dimension $ID$ depends on the number of different IDs in the training set, and $G=V=C$. 
$\textbf{b}_{s}\in {\mathbb{R}^{ID}}$ means the shared bias vector.
Subscript $I$ means image and $T$ means description.
$\textbf{P}_{\cdot }(ID)$ is the predicted probability of the right person ID, and the loss value $L_{\cdot }$ is the negative logarithm of the probability $\textbf{P}_{\cdot }(ID)$.

\textbf{Matching objective}: The hinge-based triplet matching objective has shown its strength in the task of image-text matching \cite{SCAN,VSE++}.
Referring to \cite{VSE++}, we adopt the Sum of Hinge (SH) loss ${L_{M}}$ as the matching objective:
\begin{equation}\label{equ_matchingloss_soft}
\begin{aligned}
{L_{M}} & =  \sum\limits_{\widehat T} {max [0,\alpha  - S(I,T) + S(I,\widehat T)]}  \\
& + \sum\limits_{\widehat I} {max [0,\alpha  - S(T,I) + S(T,\widehat I)]},
\end{aligned}			
\end{equation} 				
where $I$ means image and $T$ means description. 
$(I, T)$ and $(T, I)$ mean the matched image and description pairs, and $(I, \widehat T)$, $(T, \widehat I)$ indicate the mismatched pairs. 
$S( \cdot , \cdot )$ means the similarity function between two samples. 
Parameter $\alpha$ is for the margin between matched and mismatched pairs.

\textbf{Discussion.} 
The mentioned two objectives have different concerns.
The identity objective classifies the descriptions corresponding to different images while referring to the same person into the same ID category.
However, the descriptions of an image may have some kind of mismatch to other images inside the same ID category.
In other words, the identity objective is a little weak in handling the fine-grained matching.
Therefore, the identity objective is more like a loose constraint, which is fit for \textbf{initialization} in training to eliminate obvious mismatched pairs.
As for the matching objective, it is stricter because it regards the descriptions annotated for one image as negative matchings for other images that belong to the same person ID.
So the matching objective can be used to learn more accurate cross-modal relations between an image and its corresponding descriptions, which is more suitable to be exploited for \textbf{finetuning}.

\subsubsection{Training Strategy} \label{sec_train_strategy}
The step training strategy contains three steps, corresponding to the three modules in our MIA model, $i.e.$, the GC, RGA and BFM modules, respectively.

\textbf{In the first step}, we only use the identity objective to initialize the parameters related to global representations, which are annotated by number \circled{1} in Figure \ref{fig_framework}.
As explained in Section \ref{sec_objectives}, the identity objective is more like a loose constraint and fits for initialization, thus we do not fine-tune the pre-trained visual CNN but focus on training the textual path and global visual FC layer from scratch.
The overall loss function of the first step is
\begin{equation}
{L_1}{\rm{ = }}{L_I} + {L_T} \ .
\end{equation} 

\textbf{In the second step}, we aim to train the fine-grained component representations under the reference of trained global contexts, thus we additionally use the matching objective which is more suitable for accurate fine-tuning.
As annotated by numbers \circled{1} and \circled{2} in Figure \ref{fig_framework}, the parameters (including the visual CNN) are fine-tuned by the identity and matching objectives together, and the overall loss function is
\begin{equation} \label{equ_loss_step2}
{L_2}{\rm{ = }} L_1 + L_{M}^G + (L_{M}^{I - T} + L_{M}^{T - I}),
\end{equation} 
where $L_{M}^G$ means the matching objective for global representations in the GC module.
$L_{M}^{I - T}$ and $L_{M}^{T - I}$ indicate the matching objective for the two opposite directions in RGA, respectively.  

\textbf{Finally}, we fix other parameters except for the two MLPs in the BFM module for parts and phrases in training, as annotated by number \circled{3} in Figure \ref{fig_framework}.
The loss function is
\begin{equation} \label{equ_loss_step3}
{L_3}{\rm{ = }} L_{M}^{P-N} + L_{M}^{N-P}. 
\end{equation}
$L_{M}^{P-N}$ and $L_{M}^{N-P}$ are for the matching objectives in the two opposite directions in the BFM module. 

\textbf{Discussion.}
In the proposed step training strategy, the identity objective is only used to train the global contexts rather than local components, and the reason is that only global representations have tighter relevance to person ID.
Specifically, different people could have similar local components, $i.e.$, local components are not tightly relevant to person ID.
Thus it is a little inappropriate to use identity objective for classifying the local components.

\begin{table*}[]
	\centering
	\caption{Detailed configurations in experiments (S1, S2 and S3 mean separate steps in training).}
	\label{tab_conf}
	\begin{tabular}{l|ccc|ccc|ccc|ccc|ccc|cccc}
		\hline \hline
		\multirow{3}{*}{}  & \multicolumn{9}{c|}{Granularity}                                                                                                                                                & \multicolumn{6}{c|}{Training Strategy}                        & \multicolumn{4}{c}{\multirow{2}{*}{Results}} \\ \cline{2-16}
		& \multicolumn{3}{c|}{Context}                               & \multicolumn{3}{c|}{Relation}                         & \multicolumn{3}{c|}{Component}                             & \multicolumn{3}{c|}{Identity} & \multicolumn{3}{c|}{Matching} & \multicolumn{4}{c}{}                         \\ \cline{2-20} 
		& S1      & \multicolumn{1}{c}{S2} & \multicolumn{1}{c|}{S3} & S1 & \multicolumn{1}{c}{S2} & \multicolumn{1}{c|}{S3} & S1      & \multicolumn{1}{c}{S2} & \multicolumn{1}{c|}{S3} & S1         & S2        & S3   & S1       & S2       & S3      & $s_G$     & $s_R$     & $s_L$     & $s_F$     \\ \hline
		GC (G)             & $\surd$ & $\surd$                &                         &    &                        &                         &         &                        &                         & $\surd$    & $\surd$   &      &          & $\surd$  &         & $\surd$   &           &           &           \\
		GC + RGA (G)       & $\surd$ & $\surd$                &                         &    & $\surd$                &                         &         &                        &                         & $\surd$    & $\surd$   &      &          & $\surd$  &         & $\surd$   &           &           &           \\
		GC + BFM (G)       & $\surd$ & $\surd$                &                         &    &                        &                         &         & $\surd$                &                         & $\surd$    & $\surd$   &      &          & $\surd$  &         & $\surd$   &           &           &           \\
		GC + RGA + BFM (G) & $\surd$ & $\surd$                &                         &    & $\surd$                &                         &         & $\surd$                &                         & $\surd$    & $\surd$   &      &          & $\surd$  &         & $\surd$   &           &           &           \\
		MIA (G)            & $\surd$ & $\surd$                &                         &    & $\surd$                &                         &         &                        & $\surd$                 & $\surd$    & $\surd$   &      &          & $\surd$  & $\surd$ & $\surd$   &           &           &           \\ \hline
		GC + RGA (R)       & $\surd$ & $\surd$                &                         &    & $\surd$                &                         &         &                        &                         & $\surd$    & $\surd$   &      &          & $\surd$  &         &           & $\surd$   &           &           \\
		GC + RGA + BFM (R) & $\surd$ & $\surd$                &                         &    & $\surd$                &                         &         & $\surd$                &                         & $\surd$    & $\surd$   &      &          & $\surd$  &         &           & $\surd$   &           &           \\
		MIA (R)            & $\surd$ & $\surd$                &                         &    & $\surd$                &                         &         &                        & $\surd$                 & $\surd$    & $\surd$   &      &          & $\surd$  & $\surd$ &           & $\surd$   &           &           \\ \hline
		Fine (L)           &         &                        &                         &    &                        &                         & $\surd$ &                        &                         &            &           &      & $\surd$  &          &         &           &           & $\surd$   &           \\
		GC + BFM (L)       & $\surd$ & $\surd$                &                         &    &                        &                         &         & $\surd$                &                         & $\surd$    & $\surd$   &      &          & $\surd$  &         &           &           & $\surd$   &           \\
		GC + RGA + BFM (L) & $\surd$ & $\surd$                &                         &    & $\surd$                &                         &         & $\surd$                &                         & $\surd$    & $\surd$   &      &          & $\surd$  &         &           &           & $\surd$   &           \\
		MIA (L)            & $\surd$ & $\surd$                &                         &    & $\surd$                &                         &         &                        & $\surd$                 & $\surd$    & $\surd$   &      &          & $\surd$  & $\surd$ &           &           & $\surd$   &           \\ \hline
		GC                 & $\surd$ & $\surd$                &                         &    &                        &                         &         &                        &                         & $\surd$    & $\surd$   &      &          & $\surd$  &         &           &           &           & $\surd$   \\
		GC + BFM           & $\surd$ & $\surd$                &                         &    &                        &                         &         & $\surd$                &                         & $\surd$    & $\surd$   &      &          & $\surd$  &         &           &           &           & $\surd$   \\
		GC + RGA           & $\surd$ & $\surd$                &                         &    & $\surd$                &                         &         &                        &                         & $\surd$    & $\surd$   &      &          & $\surd$  &         &           &           &           & $\surd$   \\
		GC + RGA + BFM     & $\surd$ & $\surd$                &                         &    & $\surd$                &                         &         & $\surd$                &                         & $\surd$    & $\surd$   &      &          & $\surd$  &         &           &           &           & $\surd$   \\
		MIA                & $\surd$ & $\surd$                &                         &    & $\surd$                &                         &         &                        & $\surd$                 & $\surd$    & $\surd$   &      &          & $\surd$  & $\surd$ &           &           &           & $\surd$   \\ 
		\hline \hline
	\end{tabular}
\end{table*}

\section{Experiments and Analysis}
\subsection{Dataset and Protocols} \label{Protocols}
We evaluate our MIA model based on the CUHK-PEDES \cite{Person_Search_GNA-RNN} dataset, which is \textbf{currently the only one} for description-based person Re-id. 
It contains 40,206 images from 13,003 different pedestrians, and each image comes with two hand-annotated descriptions. 
The textual descriptions are generally longer than 23 words, and the vocabulary contains 9,408 unique words in total. 
We follow the same protocol in \cite{Person_Search_GNA-RNN}. 
Specifically, there are 34,054 images with 68,108 captions of 11,003 different pedestrians in the training set. 
The validation set has 3,078 images, 6,156 sentences of 1,000 pedestrians, and the testing set has 3,074 images with 6,148 captions of another 1,000 pedestrians.

We measure performance by R@K, $i.e.$, recall at K---the fraction of queries for which the correct item is retrieved in the closest K points to the query. 
Following \cite{Person_Search_GNA-RNN}, a successful search is achieved if any image of the corresponding person is among the top-K images. 
We report R@1, R@5 and R@10 criteria and their summation (Total) for all the experiments.

\subsection{Implementation Details} \label{section_implementation_details}
For an image $I$, the global visual representation $\textbf{I}$ has the dimension of $V=1024$. 
To have fair comparisons with the previous methods, we use the pre-trained VGG-16 \cite{VGG} and ResNet-50 \cite{ResNet} as visual CNN, respectively. 
Following \cite{PCB}, we resize all images to the size of 384$\times$128, and these images are randomly mirrored for image augumentation before sent to the visual encoder. 	
For VGG-16, we extract the feature maps before the first FC layer, whose size is 12$\times$4$\times$512. 
As for ResNet-50, we extract the feature maps before the average pooling layer, whose size is 24$\times$8$\times$2048.
As for the text encoding, the global textual representation $\textbf{T}$ has the dimension of $C=1024$.
Dimension of the word embeddings that are input to the GRU is $E=300$, and the textual forward and backward final hidden states $\overrightarrow {{\textbf{h}_F}}$ and $\overleftarrow {{\textbf{h}_{F'}}}$ from GRU are of $H=1024$, respectively.

For image parts, we split feature maps from visual CNN to six parts equally along the vertical direction, $i.e.$, $n=6$ in all the experiments, which has been proved to be the optimized number of parts in \cite{PCB}.
The two visual MLPs, $i.e.$, MLP-V-BFM and MLP-V-RGA, map the features of image parts from the dimension of 256 to $P=1024$. 
As for noun phrases, the output dimension of the textual MLPs, $i.e.$, MLP-T-BFM and MLP-T-RGA, are set to $N=1024$.
All the four MLPs consist of two linear mapping layers and an activation function of ReLU \cite{relu, leaky_relu} between them.

We employ the Cosine similarity as the similarity function $sim ( \cdot , \cdot )$ everywhere in our MIA model, which has been proved to be effective and widely used in the field of cross-modal retrieval.
In training, the Adam \cite{Adam} optimizer is employed to train the models with a batch size of 96. 
We start training with a learning rate of 0.001 for 10 epochs in step-1.
In step-2, the initial learning rate is set to 0.0002, and it scales down to one tenth after each 10 epochs.
After training for 15 epochs in step-2, we fix all the modules except for the two MLPs in the BFM module, $i.e.$,  MLP-V-BFM and MLP-T-BFM, for training another 5 epochs with a learning rate of 0.0002 in step-3.
The margin $\alpha$ is set to 0.2 in the matching objective $L_{M}$ in Equation \ref{equ_matchingloss_soft}.

\begin{figure*}[t]
	\centering
	\includegraphics[width=16.5cm]{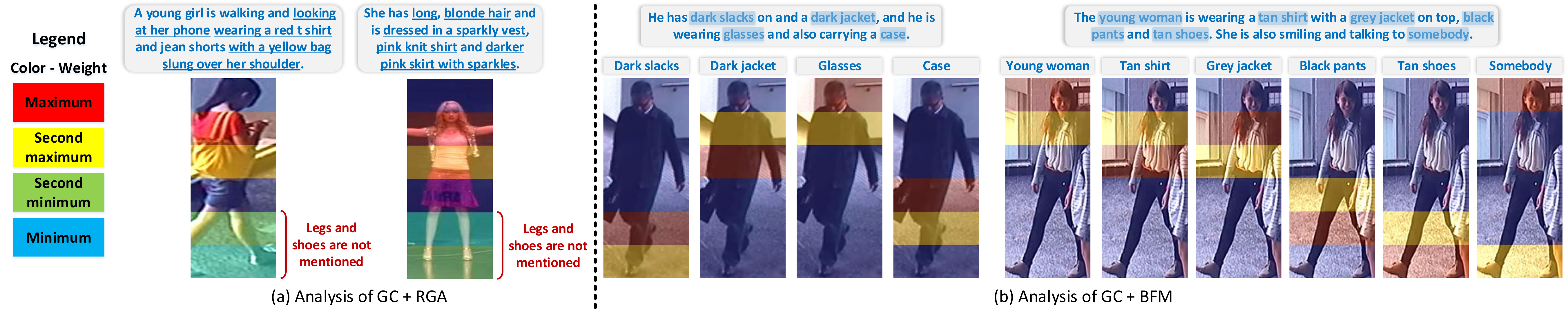}
	\vspace{-2mm}
	\caption{Visualization analysis on ablation studies of our method. (a) The effectiveness of the relation-guided attention in the RGA module. We provide the $I \to T$ direction as an example, $i.e.$, the relations between image parts and the global textual context. Red color means the part with the maximal weight after attention and yellow color means the second maximum. Green and blue colors are for the parts with the smallest weights. The underlined attributes in description are relevant to the two parts with the largest weights in image, which are the most distinguishable attributes for more accurate person identification. In contrast, the image parts that are relevant to the uninvolved components in description have the smallest weights. (b) The effectiveness of the fine-grained matching in the BFM module. The two examples are using noun phrases to attend to image parts ($P \to N$ direction). Red color means the part is the most similar one (with the maximal weight) to the phrase after the part-phrase attention, and yellow color means the second similar one. (Best viewed in colors.)} 
	\label{fig_ablation}		
\end{figure*}

\begin{figure*}[t]
	\centering
	\includegraphics[width=17cm]{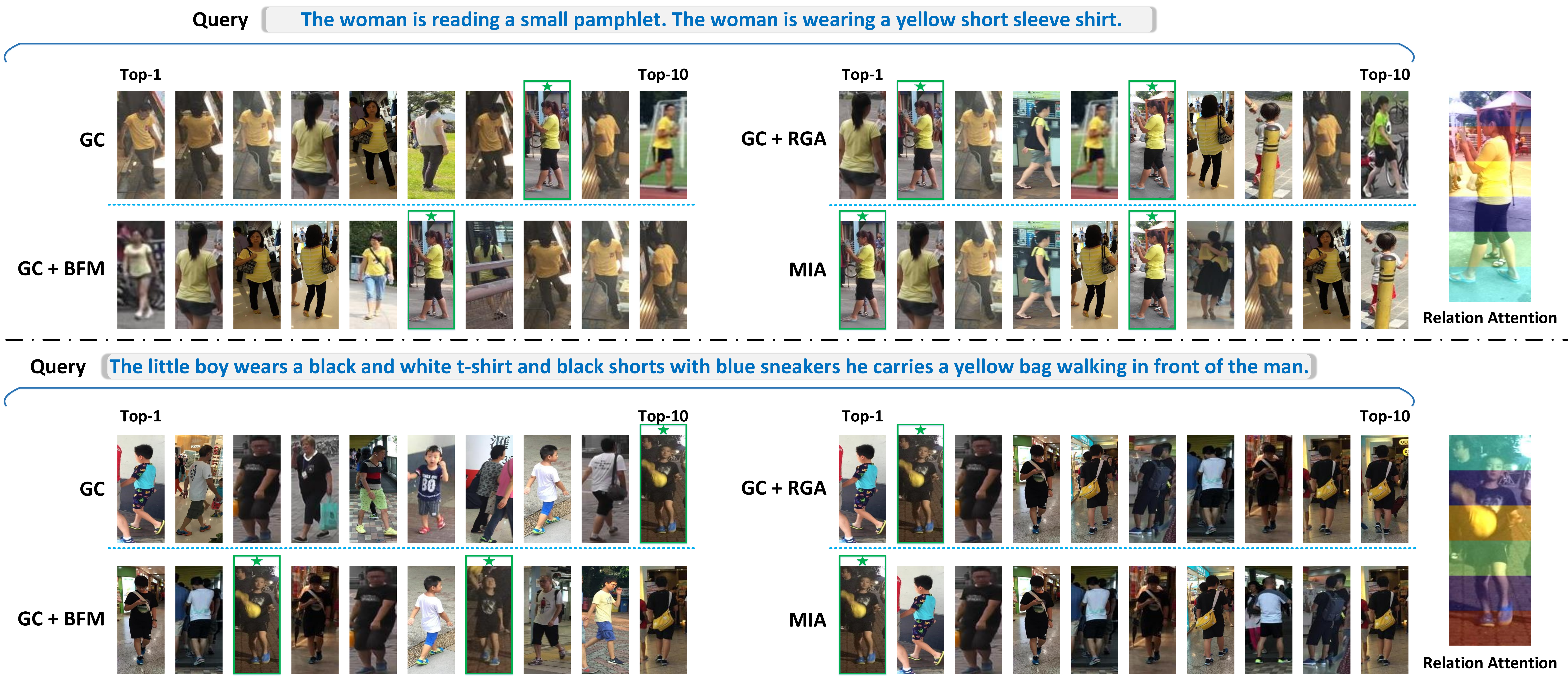}
	\vspace{-2mm}
	\caption{Comparisons of the retrieval results among different granularities. The `GC + BFM' and `GC + RGA' models outperform the `GC' model, and our `MIA' method obtains the best retrieval results by combining multiple granularities. Taking the upper one for instance, using the `yellow short sleeve shirt' can retrieve many people with a yellow shirt, but the fine-grained attribute `reading a small pamphlet' is the key semantic concept to distinguish the correct person from others. As shown by the image in right (same color meanings as in Figure \ref{fig_ablation}), the `reading a small pamphlet' part has the largest weight and the `yellow short sleeve shirt' part has the second largest weight after the relation-guided attention. On the contrary, the bottom two parts have the smallest weights because the legs and shoes are not mentioned in the query description. And the lower example can be explained similarly. (Best viewed in colors.)} 
	\label{fig_retrieval_compare}	
	\vspace{-2mm}	
\end{figure*}

\begin{table}[t]
	\footnotesize
	\centering
	\caption{Ablation study of the MIA model.}
	\label{tab_granularity}
	\begin{tabular}{l|ccc|c}
		\hline \hline
		Models          				& R@1 				& R@5 			& R@10 				& Total 			\\ \hline
		GC  							& 43.4    			& 67.4    		& 76.7     			& 167.2     		\\ 
		GC + BFM 						& 44.7    			& 69.0    		& 77.6     			& 191.3      		\\ 
		GC + RGA						& 47.2    			& 70.5    		& 79.1    			& 196.8      		\\ 
		GC + RGA + BFM 					& 46.7 				& 70.5 			& 79.1 				& 196.3 			\\ \hline
		MIA (G)							& 45.9 				& 69.1 			& 78.1 				& 193.1				\\
		MIA (R)							& 46.3 				& 69.5 			& 78.2 				& 194.0				\\ 
		MIA (L)							& 36.1 				& 59.8 			& 69.5 				& 165.4				\\
		MIA								& \textbf{48.0}   	& \textbf{70.7} & \textbf{79.3}    	& \textbf{198.0}	\\ 		 		
		\hline \hline
	\end{tabular}
	\vspace{-2mm}
\end{table}

\subsection{Evaluation of Ablation Models} \label{sec_ablation}
We carry out extensive experiments for evaluation of ablation models by taking VGG-16 as the visual CNN. 
To systematically evaluate the contributions of different model components and training configurations, we design various ablation models.
As shown in Table \ref{tab_conf}, `Context', `Relation' and `Component' are for the global-global, global-local and local-local alignments, respectively.
The S1, S2 and S3 mean separate steps in training.
The last column `Results' in Table \ref{tab_conf} indicates the retrieval results of which granularity are reported.
When a specific granularity is not trained, the corresponding parameter $\lambda$ will be set to 0 in $s_F$ in Equation \ref{equ_sim_fusion}.
For example, the entry `GC' is provided with $\lambda_1=0$ and $\lambda_2=0$. 
For the model which needs to combine multiple granularities, we set $\lambda_1=1$ and $\lambda_2=0.5$ in Equation \ref{equ_sim_fusion} to report the $s_F$.

\subsubsection{Granularities and Their Combination} \label{sec_granularity}
As our MIA model is a solution that combines multiple granularities of cross-modal similarities, we provide analysis on the effectiveness of separate granularities and their combinations, as shown in Table \ref{tab_granularity}.
Specifically, entries in the upper block show the retrieval performance when combining different granularities in training.
In the lower block, we train the whole MIA model with our step training strategy and provide the performance of different granularities individually, $i.e.$, $s_G$, $s_R$, $s_L$, and their combinations $s_F$.
The detailed analyses are as follows:

\textit{1-1)} The `GC + RGA' and `GC + BFM' both obtain performance improvements compared with the `GC', which proves that employing multi-granularity alignments can facilitate more accurate cross-modal similarity evaluation.

\textit{1-2)} The RGA module is more effective than the BFM module.
The reason is that using trained global contexts as reference in RGA tends to have the appropriate optimization direction for fine-grained feature extraction, resulting in better retrieval performance.
In contrast, directly using fine-grained components for cross-modal matching in the BFM module may suffer from the problem of ambiguous local components.

\textit{1-3)} The `GC + RGA + BFM' performs better than the `GC + BFM' but worse when compared to the `GC + RGA', which again validates the superiority of relation-guided filtering in RGA.
And it also proves that employing fine-grained component matching in BFM for training together may lead to a wrong optimizing direction and harm the retrieval accuracy.

\textit{1-4)} The `MIA (R)' is better than the `MIA (G)'.
This indicates that considering relations to adaptively highlight the significant components while filtering the uninvolved ones can obtain improvements than the global context matching.
The global contexts consider every component equally by the global pooling, ignoring their different significance and bringing some noise from the uninvolved components.

\textit{1-5)} The full model `MIA' outperforms any one of the `MIA (G)', `MIA (R)' and `MIA (L)', which indicates that combining multi-granularity image-text alignments can provide comprehensive cross-modal similarity evaluation, leading to better retrieval performance.

To show the effectiveness of different granularities more clearly, we provide visualization analysis in Figure \ref{fig_ablation}.
\textbf{In part (a)}, we illustrate some representative examples in `GC + RGA' to show the strength of the relation-guided attention in the RGA module.
Red color means that the part is the most relevant one to the textual context ($i.e.$, maximal weight in attention), which should be highlighted heavily in the visual parts aggregation guided by the global-local relations.
And the part in yellow color has the second maximum weight.
The words that related to these two marked visual parts (with the largest weights) are underlined in description, and we can find that they cover the major attributes in description, which are the key semantic concepts that could be used to distinguish the person from others. 
In contrast, the parts in green and blue colors have the smallest weights after relation-guided attention, and the reason is that these two parts are for the attributes which are not mentioned in the description. 
Taking the first image-description pair as an example, `looking at her phone', `wearing a red t-shirt' and `with a yellow bag slung over her shoulder' are the major attributes that make the person distinguishable from other people.
Our method properly focuses on the two image parts containing these key attributes and can further achieve more accurate matching.
The legs and shoes are not mentioned in the description, thus the bottom two image parts have the smallest weights.

\textbf{In part (b)}, we illustrate some representative examples in `GC + BFM' to show the effectiveness of the fine-grained matching in BFM.
Red color means the part is the most relevant one (with maximal weight) to the phrase after attention, and yellow color means the second most relevant one.
Unlike part (a), we omit the parts with the smallest weights in image, and the reason is that it is meaningless to tell which image part has the most discrepancy to the noun phrase.
In the left example in Figure \ref{fig_ablation} (b), we can find that our method can obtain correct matching for all the four noun phrases which have explicit referring attributes in image.
As for the example in the right, the first five noun phrases with explicit referring attributes are correctly referred.
But the last phrase `Somebody', which is ambiguous and involves almost all parts, cannot be accurately located to specific parts in the image.
Based on the accurate fine-grained component matching, the major difficulty in description-based Re-id, $i.e.$, the cross-modal fine-grained problem, can be alleviated.
And the retrieval performance can be further improved.

We provide comparisons of retrieval results in Figure \ref{fig_retrieval_compare}.
There are two aspects which are worth noting:
\textbf{First}, the `GC + BFM' and `GC + RGA' models both outperform the `GC' model, and `GC + RGA' is better.
Taking the upper one for example, we visualize the relation attention in RGA and find that our model focuses on `reading a small pamphlet' (in red with the maximal weight) and `wearing a yellow short sleeve shirt' (in yellow with the second maximal weight), which are the most distinguishable attributes for accurate retrieval.
Between these two attributes, using `yellow short sleeve shirt' can retrieve many people with a yellow shirt, but `reading a small pamphlet' is the key attribute to distinguish the correct person from others.
Therefore, the part in red is the most distinguishable one, $i.e.$, having maximal weight in our relation-guided attention filtering.
\textbf{Second}, by combining the RGA and BFM modules, the MIA model can obtain the best retrieval results than either RGA or BFM individually, validating the effectiveness of combining multi-granularity alignments in cross-modal similarity evaluation.

\begin{table}[t]
	\footnotesize
	\centering
	\caption{Analysis of the proposed step training strategy.}
	\label{tab_step_train}
	\begin{tabular}{lcccc}
		\hline \hline
		\multicolumn{1}{l|}{Models}                      	& R@1  				& R@5  			& \multicolumn{1}{c|}{R@10} 			& Total 			\\ \hline
		\multicolumn{5}{c}{$s_G$}                                                       																	\\ \hline
		\multicolumn{1}{l|}{GC (G)}        					& 43.4  			& 67.4  		& \multicolumn{1}{c|}{76.7} 			& 187.8   			\\ 
		\multicolumn{1}{l|}{GC + BFM (G)}        			& 42.8  			& 66.2  		& \multicolumn{1}{c|}{75.7} 			& 184.7   			\\ 
		\multicolumn{1}{l|}{GC + RGA (G)}        			& \textbf{45.9} 	& \textbf{69.1} & \multicolumn{1}{c|}{\textbf{78.1}} 	& \textbf{193.1}   	\\
		\multicolumn{1}{l|}{GC + RGA + BFM (G)}        		& 44.1 				& 68.2  		& \multicolumn{1}{c|}{77.0} 			& 189.3   			\\
		\multicolumn{1}{l|}{MIA (G)}        				& \textbf{45.9} 	& \textbf{69.1} & \multicolumn{1}{c|}{\textbf{78.1}} 	& \textbf{193.1}   	\\ \hline
		\multicolumn{5}{c}{$s_R$}                                                                  															\\ \hline
		\multicolumn{1}{l|}{GC + RGA (R)}        			& \textbf{46.3} 	& \textbf{69.5} & \multicolumn{1}{c|}{\textbf{78.2}} 	& \textbf{194.0}  	\\
		\multicolumn{1}{l|}{GC + RGA + BFM  (R)}        	& 44.2 				& 68.9  		& \multicolumn{1}{c|}{78.2} 			& 191.3  			\\ 
		\multicolumn{1}{l|}{MIA (R)}        				& \textbf{46.3} 	& \textbf{69.5} & \multicolumn{1}{c|}{\textbf{78.2}} 	& \textbf{194.0}  	\\ \hline
		\multicolumn{5}{c}{$s_L$}                                                                  															\\ \hline
		\multicolumn{1}{l|}{Fine (L)}        				& 32.0 				& 55.5 			& \multicolumn{1}{c|}{66.0} 			& 153.5				\\
		\multicolumn{1}{l|}{GC + BFM (L)}        			& 34.0 				& 57.4 			& \multicolumn{1}{c|}{67.9} 			& 159.3				\\
		\multicolumn{1}{l|}{GC + RGA + BFM (L)}        		& 34.7 				& 58.6 			& \multicolumn{1}{c|}{69.4} 			& 162.7 			\\
		\multicolumn{1}{l|}{MIA (L)} 						& \textbf{36.1} 	& \textbf{59.8} & \multicolumn{1}{c|}{\textbf{69.5}} 	& \textbf{165.4} 	\\ \hline
		
		\multicolumn{5}{c}{$s_F$}                                                                  															\\ \hline
		\multicolumn{1}{l|}{GC + RGA + BFM}        			& 46.7 				& 70.5 			& \multicolumn{1}{c|}{79.1} 			& 196.3 			\\
		\multicolumn{1}{l|}{MIA} 							& \textbf{48.0} 	& \textbf{70.7} & \multicolumn{1}{c|}{\textbf{79.3}} 	& \textbf{198.0} 	\\ 
		\hline \hline
	\end{tabular}
	\vspace{-3mm}
\end{table}
\begin{table}[t]
	\centering
	\caption{Analysis of employing different objectives in the two steps (GC + RGA results).}
	\label{tab_steps}
	\begin{tabular}{l|l|lll|l}
		\hline \hline
		\multicolumn{2}{c|}{Methods} & \multicolumn{1}{c}{\multirow{2}{*}{R@1}} & \multicolumn{1}{c}{\multirow{2}{*}{R@5}} & \multicolumn{1}{c|}{\multirow{2}{*}{R@10}} 
		& \multicolumn{1}{c}{\multirow{2}{*}{Total}} \\ \cline{1-2}
		Step-1        & Step-2       & \multicolumn{1}{c}{}                     & \multicolumn{1}{c}{}                     & \multicolumn{1}{c|}{}                      
		& \multicolumn{1}{c}{}                       \\ \hline
		Mat           & Mat          & 40.6                                     & 64.9                                     & 74.3                                           
		& 179.8                                      \\ 
		Mat           & Id + Mat     & 42.1                                     & 66.4                                     & 75.8                                           
		& 184.3                                      \\ 
		Id            & Id + Mat     & \textbf{47.2}                            & \textbf{70.5}                            & \textbf{79.1}                              
		& \textbf{196.8}                             \\
		\hline \hline
	\end{tabular}
	\vspace{-2mm}
\end{table}

\subsubsection{Step Training Strategy Analysis}

In Table \ref{tab_step_train}, we provide the detailed results in different granularities individually, $i.e.$, $s_G$, $s_R$, $s_L$ and the final similarity $s_F$, to carry out analysis on the proposed step training strategy.

\textit{2-1)} 
For the global context similarity $s_G$, we can find that directly employing the BFM module for training together tends to harm the performance based on the comparisons between `GC + BFM (G)' and `GC (G)' as well as `GC + RGA + BFM (G)' and `GC + RGA (G)'.
The reason is due to ambiguities of local components, which may lead to a wrong optimizing direction and influence the quality of feature extractions.

\textit{2-2)} Compared between `MIA (G)' and `GC + RGA + BFM (G)', we can find that our step training strategy can alleviate the problem of ambiguous local component extraction, and keep the $s_G$ from decreasing.

\textit{2-3)} The `MIA (G)' has the same results as the `GC + RGA (G)'.
The reason is that we fix the parameters related to the GC and RGA modules but focus on training the two MLPs (MLP-V-BFM and MLP-T-BFM) in step-3.
So the parameters related to the GC and RGA modules remain unchanged. 

\textit{2-4)} Models in the $s_R$ block have the similar behaviors as in the $s_G$ block, and the reasons are the same as well.

\textit{2-5)} In the block of $s_L$, we can find that the performance of $s_L$ improves gradually.
From `Fine (L)' to `MIA (L)', our step training strategy introduces the GC, RGA and BFM modules step-by-step with appropriate objectives in respective training steps. 
This will enhance the shared backbone networks and further improve the fine-grained component feature extraction, leading to a better fine-grained similarity $s_L$.

\textit{2-6)} As for the last block of $s_F$, our step training (`MIA') outperforms the together training (`GC + RGA + BFM') by 1.3\% in terms of R@1.
In addition, compared with the `GC + RGA' in Table \ref{tab_granularity}, `MIA' further obtains 0.8\% improvements in terms of R@1 despite the problem of ambiguities from fine-grained component extraction.
The reason is that we first use GC and RGA to train the backbone modules for better feature extraction in the step-1 and step-2, avoiding the influence of inappropriate optimizing direction in BFM.
After that, we fix the backbone modules and only train the two MLPs in BFM, which provide adaptation for more accurate $s_L$.
And this can results in a better $s_F$ finally by similarity combination. 

\subsubsection{Objective Analysis} \label{sec_objective_analysis}
Some varieties of employing different objectives in separate training steps are carried out and the results are in Table \ref{tab_steps} (Id is for identity objective and Mat means matching objective).
We choose the `GC + RGA' model for comparison, because the BFM module only contains the MLP-V-BFM and MLP-T-BFM to be trained in step-3, and the gradients are not back-propagated through other modules.
So the performance of `GC + RGA' remains unchanged.

\textit{3-1)}
Compared between the first two entries, we can find that using the identity objective and matching objective together in step-2 obtains better results than only using the matching objective.
The identity objective fits for the goal of person identification, and the matching objective is effective and commonly used in the cross-modal image-sentence matching.
Therefore, combining these two objectives together can further improve the representations in training and obtain better retrieval performance.

\textit{3-2)}
Considering the last two entries, using the identity objective in step-1 obtains better performance than the matching objective.
It is more suitable to use the identity objective that is consistent with the testing protocol \cite{Person_Search_GNA-RNN} to initialize the parameters from scratch.
Specifically, a successful search is achieved if any image of the corresponding person is among the top images. 
In contrast, the matching objective is too strict to provide an appropriate initial training orientation, because it regards the descriptions annotated for one image as negative matchings for other images that belong to the same person identity.

\begin{table}[t]
	\centering
	\caption{Comparisons between combining step-1 and step-2 and our step training strategy (GC + RGA results).}
	\label{tab_combiningsteps}
	\begin{tabular}{l|ccc|c}
		\hline \hline
		Methods                     & R@1 			& R@5 			& R@10 			& Total 			\\ \hline
		Combining step-1 \& step-2 & 39.7    		& 64.0    		& 73.9     		& 177.6      		\\ 
		Step training strategy      & \textbf{47.2} & \textbf{70.5} & \textbf{79.1} & \textbf{196.8}   \\ 
		\hline \hline
	\end{tabular}
	\vspace{-2mm}
\end{table}
\begin{table}[t]
	\footnotesize
	\centering
	\caption{Comparisons between using all words and only noun phrases (MIA results).}
	\label{tab_all_words}
	\begin{tabular}{l|ccc|c}
		\hline \hline
		Methods      		& R@1 				& R@5 				& R@10 				& Total 			\\ \hline
		All words    		& 44.2     			& 66.3   			& 75.9         		& 186.4       		\\
		Noun phrases 		& \textbf{48.0}    	& \textbf{70.7}     & \textbf{79.3}     & \textbf{198.0} 	\\
		\hline \hline
	\end{tabular}
	\vspace{-2mm}
\end{table}

\subsubsection{Step Combination}
We also combine the step-1 and step-2 together in training (using a learning rate of 0.0002 to train the whole model, including the visual CNN), and the results of the `GC + RGA' model are shown in Table \ref{tab_combiningsteps}. 
We can find its performance is much worse than our step training strategy.
This validates the importance of only using the identity objective for parameters initialization in step-1, $i.e.$, initializing the textual paths and global visual FC layer from scratch.

\subsubsection{All Words vs Noun Phrases}

To validate the superiority of using noun phrases rather than all words in description, we provide the results in Table \ref{tab_all_words}.
The two entries both employ the `MIA' model in Table \ref{tab_conf} with $\lambda_1=1$ and $\lambda_2=0.5$ for similarity combination.
We can find that the proposed MIA model obtains better performance with only noun phrases rather than all words.
The reason is that some words may not have the explicit corresponding components in the image, and these words will introduce some noise to the cross-modal fine-grained matching.
Besides, a single image part may correspond to multiple separate words in description, while a single word is not able to completely describe the image part.
	
\begin{table}[t]
	\footnotesize
	\centering
	\caption{Comparisons with the state-of-the-art methods.}
	\label{tab_results}
	\begin{tabular}{lcccc}
		\hline \hline
		\multicolumn{1}{l|}{Methods}            								& R@1   			& R@5   			& \multicolumn{1}{c|}{R@10}  			& Total 			\\ \hline 
		\multicolumn{5}{c}{VGG-16 as visual CNN}                                            																				\\ \hline
		\multicolumn{1}{l|}{CNN-RNN (2016) \cite{CNN-RNN}}     					& 8.07  			& -     			& \multicolumn{1}{c|}{32.47} 			& -      			\\
		\multicolumn{1}{l|}{Neural Talk (2015) \cite{Neural_Talk} } 			& 13.66 			& -     			& \multicolumn{1}{c|}{41.72} 			& -     			\\
		\multicolumn{1}{l|}{GNA-RNN (2017) \cite{Person_Search_GNA-RNN}	}     	& 19.05 			& -     			& \multicolumn{1}{c|}{53.64} 			& -      			\\
		\multicolumn{1}{l|}{IATVM (2017) \cite{IATVM} }       					& 25.94 			& -     			& \multicolumn{1}{c|}{60.48} 			& -      			\\
		\multicolumn{1}{l|}{PWM-ATH (2018) \cite{PWM-ATH} }     				& 27.14 			& 49.45 			& \multicolumn{1}{c|}{61.02} 			& 137.61    		\\
		\multicolumn{1}{l|}{Dual Path (2017) \cite{Dual_Path}}   				& 32.15 			& 54.42 			& \multicolumn{1}{c|}{64.30} 			& 150.87    		\\ \hline	
		\multicolumn{1}{l|}{MIA ($\lambda_1=1$, $\lambda_2=0.5$)}         		& \textbf{48.00} 	& \textbf{70.70}  	& \multicolumn{1}{c|}{\textbf{79.30}}	& \textbf{198.00}	\\ \hline 
		\multicolumn{5}{c}{ResNet-50 as visual CNN}                                         																				\\ \hline		
		\multicolumn{1}{l|}{Dual Path (2017) \cite{Dual_Path}}   				& 44.40 			& 66.26 			& \multicolumn{1}{c|}{75.07} 			& 185.73      		\\ 
		\multicolumn{1}{l|}{GLIA (2018) \cite{GLIA} }        					& 43.58 			& 66.93 			& \multicolumn{1}{c|}{76.26} 			& 186.77      		\\ \hline 
		\multicolumn{1}{l|}{MIA ($\lambda_1=1$, $\lambda_2=0.3$)}         		& \textbf{53.10}   	& \textbf{75.00}   	& \multicolumn{1}{c|}{\textbf{82.90}}  	& \textbf{211.00}   \\ 
		\hline \hline
	\end{tabular}
\vspace{-2mm}	
\end{table}
\begin{table}[t]
	\centering
	\caption{Analysis of hyper-parameter $\lambda_1$ (MIA results, $\lambda_2=0$).}
	\label{tab_para1}
	\begin{tabular}{l|ccc|c}
		\hline \hline
		$\lambda_1$   	& R@1 			& R@5 				& R@10 				& Total 			\\ \hline
		0.1 			& 46.2    		& 69.8    			& 77.9     			& 193.9    			\\
		0.3 			& 46.6    		& 70.0    			& 78.7    			& 195.3      		\\
		0.5 			& 46.9 			& 70.4 				& 78.9				& 196.2      		\\
		0.7 			& 47.0    		& 70.4    			& 79.1    			& 196.5     		\\
		0.9 			& 47.1    		& 70.4    			& 79.1    			& 196.6      		\\
		1.0 			& 47.2 			& 70.5    			& 79.1   			& 196.8				\\  
		1.1 			& 47.2     		& \textbf{70.6}    	& \textbf{79.3}     & 197.1    			\\
		1.3 			& 47.3     		& 70.5   			& \textbf{79.3}    	& 197.1      		\\
		1.5 			& \textbf{47.6} & 70.5 				& 79.2				& \textbf{197.3}    \\
		1.7 			& 47.5     		& 70.5    			& \textbf{79.3}    	& \textbf{197.3}    \\
		1.9 			& 47.5    		& 70.5    			& 79.2  			& 197.2    			\\
		2.0 			& 47.4 			& 70.5     			& 79.1  			& 197.0				\\  
		\hline \hline
	\end{tabular}
	\vspace{-2mm}
\end{table}

\vspace{-2mm}
\subsection{Comparison with Other State-of-the-art Methods}
The comparisons with other state-of-the-art methods are shown in Table \ref{tab_results}, and the best results are in bold. 
All the methods follow the same protocol in \cite{Person_Search_GNA-RNN} for fair comparison, as stated in Section \ref{Protocols}.
These methods can be divided into two categories by visual CNNs, $i.e.$, VGG-16 and ResNet-50. 	
Using VGG-16, we achieve significant 15.85\% improvements in terms of R@1 compared with the best Dual-Path model \cite{Dual_Path}. 
PWM-ATH \cite{PWM-ATH} employs fine-grained patch-word matching, but does not have the global-local relations for supervising the learning of fine-grained component features or filtering the uninvolved components in image-description pair.
And the multi-granularity similarity combination is not considered, either.
Our MIA model achieves 20.86\% increase compared with the PWM-ATH in terms of R@1, validating the great advantage of the global-local relations and multi-granularity alignments.
As for using the stronger ResNet-50 as visual CNN, our MIA model beats the best Dual-Path \cite{Dual_Path} model by 8.70\% in terms of R@1. 
The GLIA \cite{GLIA} introduces the patch-phrase matching, but only focuses on improving the visual representations by phrase reconstruction in local association.
Both using ResNet-50, our MIA method outperforms the GLIA by 9.52\% in terms of R@1, which again proves the superiority of our method.

\vspace{-2mm}
\subsection{Effect of the Hyper-parameters}	
In Equation \ref{equ_sim_fusion}, there are two balancing hyper-parameters $\lambda_1$ and $\lambda_2$ to adjust the proportions of separate similarities coming from different granularities.
We carry out experiments to analyze the effect of these hyper-parameters, and the detailed results are in Tables \ref{tab_para1} and \ref{tab_para2}.
Specifically, we employ the `MIA' model in Table \ref{tab_conf}, and adjust the $\lambda_1$ or $\lambda_2$ when fixing the other one to figure out their effect individually.
We can find that the performance is improved when $\lambda_1$ increases, and reaches the peak at about $\lambda_1=1.5$.
After that, it goes down a little when $\lambda_1$ keeps increasing.
As for the hyper-parameter $\lambda_2$, the performance increases before around 0.7, then a little decrease will be seen as $\lambda_2$ continues to increase.

\begin{table}[t]
	\centering
	\caption{Analysis of hyper-parameter $\lambda_2$ (MIA results, $\lambda_1=0$).}
	\label{tab_para2}
	\begin{tabular}{l|ccc|c}
		\hline \hline
		$\lambda_2$ 	& R@1 			& R@5 				& R@10 				& Total 			\\ \hline
		0.1 			& 46.5     		& 69.8    			& 78.4     			& 194.7      		\\
		0.3 			& 47.2     		& 70.1    			& 78.9     			& 196.2     		\\
		0.5 			& 47.6 			& \textbf{70.2} 	& 79.2 				& 197.0       		\\
		0.7 			& \textbf{47.8} & 70.1    			& 79.3     			& \textbf{197.2}   	\\
		0.9 			& 47.7    		& 70.0    			& \textbf{79.4}     & 197.1     		\\ 
		1.0				& 47.5     		& 69.9   			& 79.2    			& 196.6    			\\ 
		1.1 			& 47.4    		& 69.8    			& 79.1     			& 196.3    			\\
		1.3 			& 46.7     		& 69.7   			& 78.8    			& 195.2      		\\
		1.5 			& 46.3  		& 69.5 				& 78.5				& 194.3    			\\
		1.7 			& 45.9     		& 69.1    			& 78.3    			& 193.3     		\\
		1.9 			& 45.5    		& 68.6   			& 78.1   			& 192.2     		\\
		2.0 			& 45.3 			& 68.5    			& 78.1  			& 191.9				\\
		\hline \hline
	\end{tabular}
	\vspace{-2mm}
\end{table}

\subsection{Failure Cases Analysis}

\begin{figure}[t]
	\centering
	\includegraphics[width=9cm]{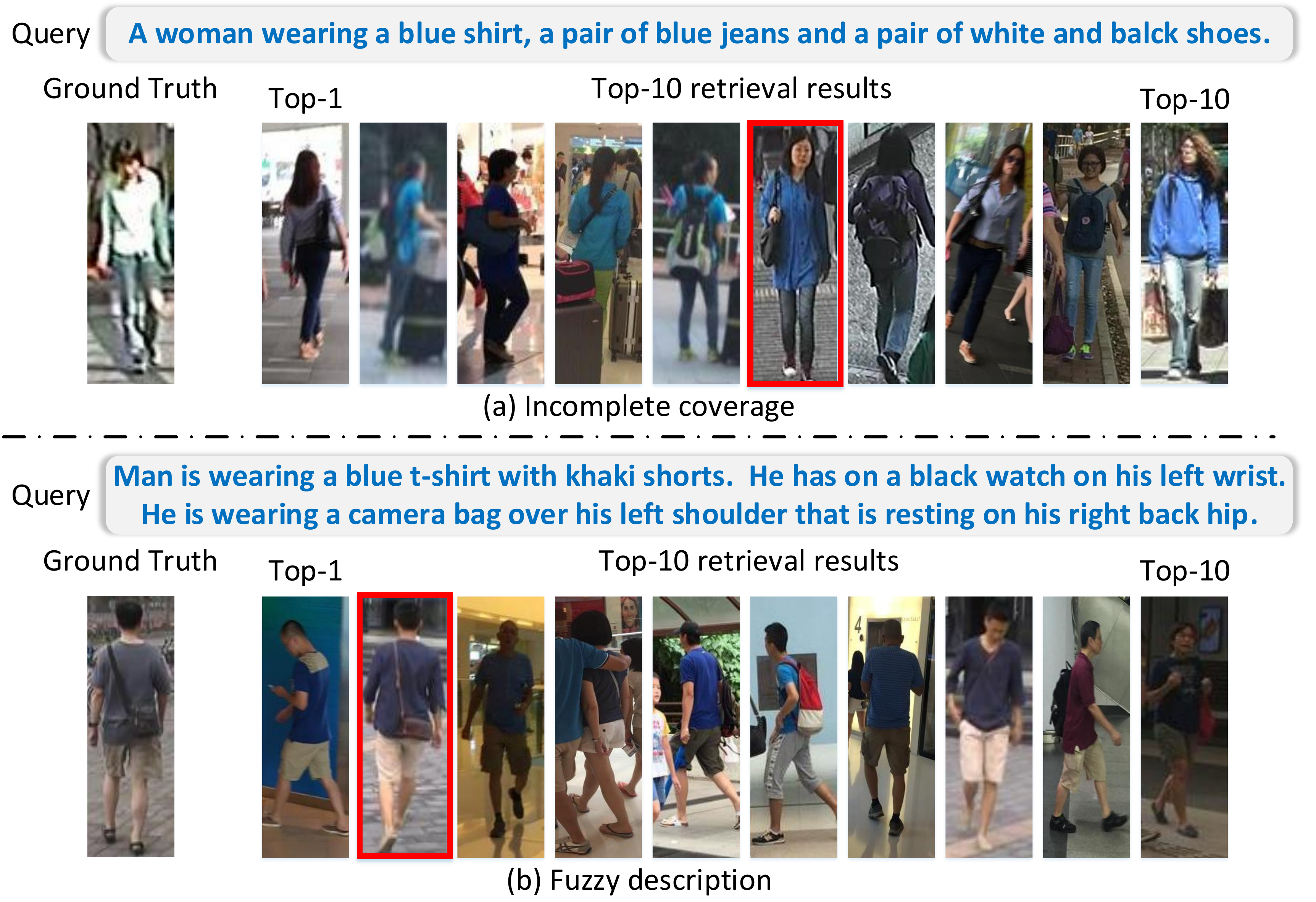}	
	\caption{Failure cases analysis. We provide some failure cases that our MIA model cannot retrieve the ground truth image within the top-10 results. These failure circumstances can be roughly divided into two different scenarios: (a) incomplete coverage and (b) fuzzy description. (Best viewed in colors.)} 
	\label{fig_failure}	
	\vspace{-2mm}	
\end{figure}

Although our MIA model has achieved significant improvements than other state-of-the-art methods, there are still some circumstances that cannot be well handled.
Some failure cases are shown in Figure \ref{fig_failure}.
Specifically, we select the cases where the true matching pedestrian images are out of the top-10 retrieval results, and they can be roughly divided into two different scenarios.
\textbf{(a) Incomplete coverage.} The query description does match some images which are not the ground-truth, but it cannot cover all the attributes in these images.
Regarding these images as correct matchings will make mistakes.
As shown in Figure \ref{fig_failure} (a), the description does match the pedestrian image in red box, but cannot cover the attribute `black bag in right shoulder', which is the key attribute to be distinguished.
And the ground truth image is mistakenly placed out of the top-10 retrieval results. 
\textbf{(b) Fuzzy description.} The sentence may contain some fuzzy attributes, which will harm the accurate retrieval.
For example in Figure \ref{fig_failure} (b), the man's t-shirt is a little fuzzy between gray color and blue color, and the description finally regards it in blue color.
On the contrary, the trained model considers the t-shirt is in gray color and places the ground truth image out of the top-10 retrieval results.
The pedestrian image in red box is more likely to have blue t-shirt in the view of the trained model which has seen thousands of images in training.		

To address these problems, we consider making more improvements to our method in the future work.
To be more specific, we may refine the image partition approach, and make the number of image parts relevant to the attributes of a pedestrian.
After that, the constraints between the number of image parts and the number of noun phrases can be exploited to alleviate the problem of incomplete coverage.
As for the fuzzy description, data cleaning and image quality enhancement ($e.g.,$ super-resolution approaches) methods are likely to solve this problem. 

\vspace{-2mm}
\section{Conclusion}
In this paper, we have proposed an end-to-end \textbf{Multi-granularity Image-text Alignments} (MIA) model for the description-based person re-identification.
The MIA model addresses the cross-modal fine-grained problem based on three different granularities hierarchically, \textit{i.e.}, global-global, global-local and local-local alignments.
Specifically, the global-global alignment in the Global Contrast (GC) module is for matching the global contexts of images and descriptions.
The global-local alignment exploits the relations between local components and global contexts to highlight the distinguishable components while eliminating the uninvolved ones adaptively in the Relation-guided Global-local Alignment (RGA) module.
And the visual human parts and noun phrases are matched in the Bi-directional Fine-grained Matching (BFM) module for the local-local alignment.
For better training the combination of multiple granularities, an effective step training strategy has been proposed to train these granularities step-by-step.
Extensive experiments and analysis have been provided to validate the effectiveness of our MIA model and the step training strategy.
This work has obtained the state-of-the-art performance on the CUHK-PEDES \cite{Person_Search_GNA-RNN} dataset, outperforming other previous methods significantly.

\ifCLASSOPTIONcaptionsoff
  \newpage
\fi

\bibliographystyle{ieee}

\bibliography{ref_tip}

\end{document}